%% file: main.tex
\documentclass[10pt,twocolumn,letterpaper]{article}

\usepackage{cvpr}
\usepackage{mathtools}
\usepackage{titletoc}
\def\paperID{6127}
\def\confName{CVPR}
\def\confYear{2026}

\input{preamble}

\title{\ourframework: From One Image to a 3D Audio-Visual Scene
}

\author{Derong Jin\thanks{Equal contribution.} \quad Xiyi Chen\footnotemark[1] \quad Ming C. Lin \quad Ruohan Gao\\[2pt]
University of Maryland, College Park\\[2pt]
}

\begin{document}
\maketitle

\begin{abstract}

Tremendous progress in visual scene generation now turns a single image into an explorable 3D world, yet immersion remains incomplete without sound. We introduce \ourtask, the task of generating a 3D audio-visual scene from a single image, and present \ourframework, the first framework to tackle this challenge. From one image, our pipeline outpaints a 360° panorama, lifts it into a navigable 3D scene, places language-guided sound anchors, and renders ambisonics for point, areal, and ambient sources, yielding spatial audio aligned with scene geometry and semantics. Quantitative evaluations on a newly curated real-world dataset and a controlled user study confirm the effectiveness of our approach. Beyond free-viewpoint audio-visual rendering, we also demonstrate applications to one-shot acoustic learning and audio-visual spatial source separation. Project website: {\url{https://humathe.github.io/sonoworld/}}
\end{abstract}

\input{sections/1_intro}
\input{sections/2_related}
\input{sections/3_task_formulation}
\input{sections/4_approach}
\input{sections/5_results}

\input{sections/6_conclusion}

\newpage
{
    \small
    \bibliographystyle{ieeenat_fullname}
    \bibliography{main}
}

\clearpage
\appendix
\input{supp}

\end{document}

%% file: preamble.tex
\usepackage{multirow}
\usepackage{siunitx}
\usepackage{algorithm}
\usepackage[noend]{algpseudocode}
\definecolor{cvprblue}{rgb}{0.21,0.49,0.74}
\usepackage[pagebackref,breaklinks,colorlinks,allcolors=cvprblue]{hyperref}
\definecolor{royalblue(traditional)}{rgb}{0.0, 0.14, 0.4}
\definecolor{darkgoldenrod}{rgb}{0.72, 0.53, 0.04}
\definecolor{maroon}{rgb}{0.5, 0.0, 0.0}
\definecolor{blazeorange}{rgb}{1.0, 0.4, 0.0}
\definecolor{marylandred}{rgb}{0.89, 0.01, 0.2}

\definecolor{vblue}{rgb}{0.12941176, 0.37254902, 0.60392157}
\definecolor{sgreen}{rgb}{0.23137255, 0.49019608, 0.137254}
\definecolor{aorange}{rgb}{0.75294118, 0.30980392, 0.08235294}
\definecolor{ipurple}{rgb}{0.47058824, 0.1254902 , 0.43137255}

\definecolor{nicegreen}{rgb}{0.1, 0.6, 0.2}

\def\ourtask{\textsc{Image2AVScene}\xspace}
\def\ourdataset{\textsc{SonoScene360}\xspace}
\def\ourframework{\textsc{SonoWorld}\xspace}

%% file: sections/1_intro.tex
\vspace*{-1em}
\section{Introduction}
\label{sec:intro}

\begin{figure}[t]
    \centering
    \includegraphics[width=1\linewidth]{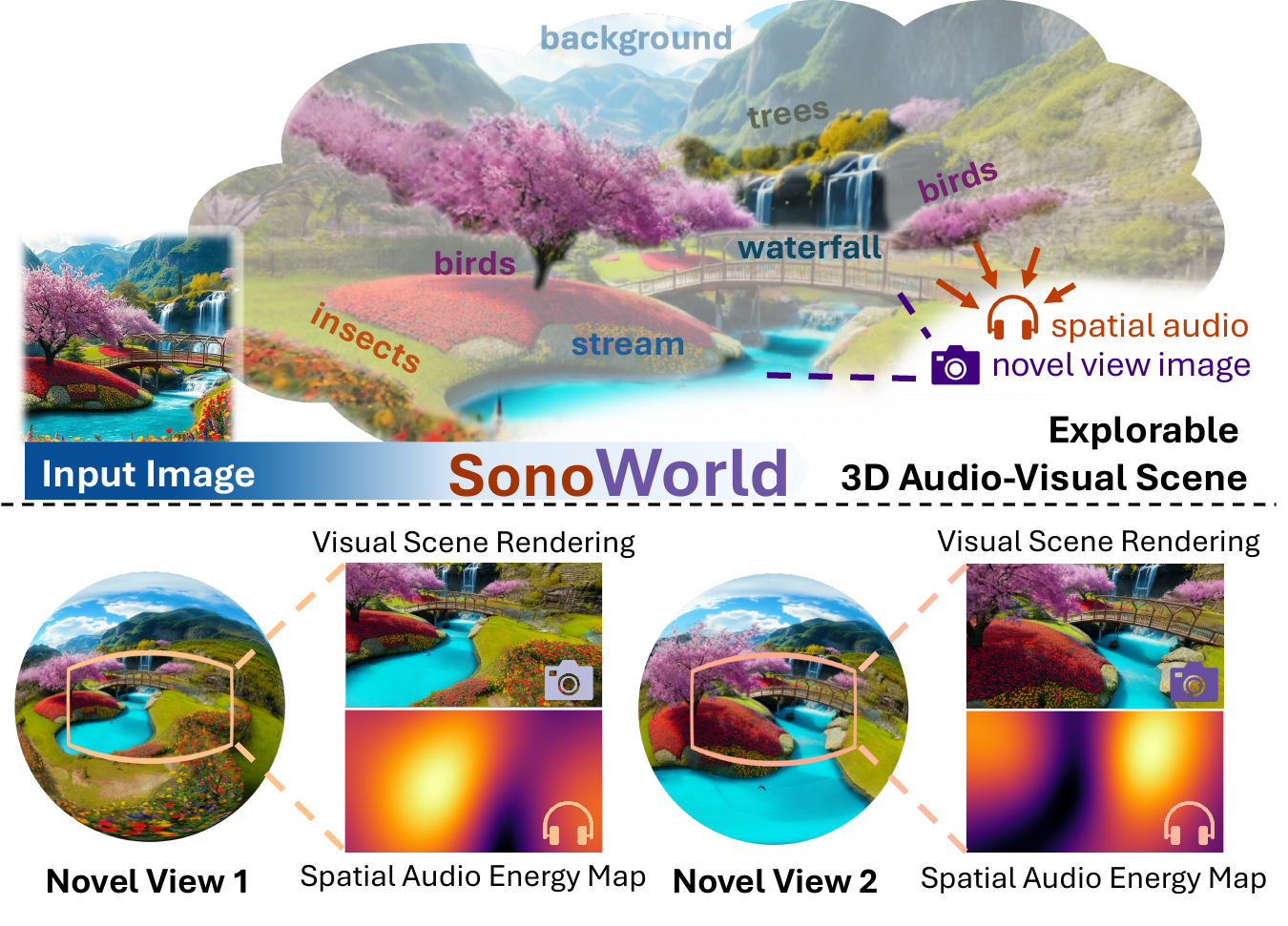}
    \vspace*{-2em}
    \caption{From one image, \ourframework generates an explorable 3D audio-visual scene, where you can navigate to novel views and locations, while listening to spatial audio aligned with scene semantics and the 3D locations of heterogeneous sound sources.
    }
    \vspace*{-1.5em}    
    \label{fig:teaser}
\end{figure}

The past few years have seen rapid progress in visual scene generation~\cite{yang2024cogvideox, zhou2025stable, Mark2025TrajectoryCrafterRC, Yu2024WonderWorldI3}. Building on recent advances in 3D scene generation~\cite{Yu2023WonderJourneyGF,Yang2025LayerPano3DL3,hunyuanworld2025tencent,Ren2025GEN3C3W,Mark2025TrajectoryCrafterRC,worldlabs_marble}, today's systems can generate photorealistic 3D worlds from a single 2D image. For example, in Fig.~\ref{fig:teaser}, one photo of a garden becomes an explorable 3D scene: you step onto a wooden bridge spanning a turquoise stream, peer beneath the arch as water tumbles through, and shift viewpoints to inspect the tiered waterfalls and overhanging cherry blossoms. Such capabilities enable compelling applications in VR/AR, content creation, and robotics. Yet, these models share a striking limitation: they produce \emph{silent} worlds you can walk through, but not \emph{listen} to.

Immersion in the real world is inherently multisensory. In the same scene, sound is crucial  for perceiving and understanding space: the waterfalls should thunder from upstream and swell as you approach; birds should chirp and leaves rustle from the canopy; insects should buzz near the flowerbeds and shift with head turns. Without these semantically meaningful sounds, together with the directional and distance cues, the world may look convincing yet remains perceptually incomplete.

To realize this vision, we introduce \ourtask, a new task that aims to generate an explorable 3D audio-visual scene from a single RGB image. Given one image of a scene, the goal is to generate (i) a navigable 3D visual scene and (ii) a spatial sound field that is semantically and geometrically aligned with the visual content. This enables users not only to look around the generated 3D scene, but also \emph{listen} from any location and viewpoint. 

Jointly generating a coherent 3D scene and its spatial sound field from a single image is challenging. \emph{First,} unlike traditional audio synthesis~\cite{audiogen,mmaudio,audioldm,owens2016visually,chen2017} that often targets isolated objects or events, scene-level audio generation must compose heterogeneous source types and scales: point sources (\eg, a chirping bird), areal sources (\eg, a flowing river), and ambient soundscapes (\eg, forest insects and wind). Each behaves differently over time and distance, and must remain coherent as the listener moves. \emph{Second,} the system requires holistic \emph{semantic} scene understanding to infer what is likely to make sound, how it sounds, and how loud all from the visual context: waterfalls roar and fluctuate, birds call intermittently, insects form a high-frequency bed near flowers and trees---while silent objects (\eg, a wooden bridge) should remain silent. \emph{Third,} all generated sounds must be grounded to plausible 3D locations and spatial extents inferred from the image, and rendered with perceptually realistic \emph{spatial} effect (\eg, direction of arrival and distance-dependent attenuation).

We propose \ourframework, a \emph{training-free} framework to systematically address these challenges. Given a single image as input, we first reproject it to an equirectangular panorama with elevation correction and outpaint a complete 360° view. We then lift this panorama into a navigable 3D representation (3D Gaussian splats \cite{3dgs}) to obtain dense, view-consistent geometry. To bridge vision and sound, we perform 360° semantic grounding: a language-guided set of sounding categories seeds open-vocabulary instance discovery on tiled perspective views; these instances are reconciled with panoramic mask proposals and back-projected into 3D to produce instance/region anchors for sound. Finally, a spatial audio encoder converts these anchors into ambisonics coefficients, supporting point-like and areal sources as well as ambient sound, yielding geometry-aware spatialization consistent with the scene’s 3D structure and semantics. 

To facilitate evaluation of this new task, we contribute an evaluation dataset of 68 clips that contains both panorama videos and ambisonics recordings across six diverse real-world scenes, and we define a suite of metrics assessing both the semantic fidelity and spatial accuracy of the generation outputs. Across these metrics and a user study, our method consistently outperforms a series of baselines that generate spatial audio from visual input. Beyond free-viewpoint audio-visual rendering and exploration in the generated 3D audio-visual scene, we also show evidence that our framework extends to two additional 3D audio-visual learning tasks: one-shot room acoustic learning and audio-visual spatial source separation.

In summary, our key contributions are:
\vspace*{-0.5em}
\begin{itemize}[itemsep=1pt, topsep=2pt,leftmargin=12pt]
\item We introduce \ourtask, a novel task to generate an interactive 3D visual scene together with a spatial sound field that is semantically and geometrically grounded to the visual context, along with the first effective framework, \ourframework, to tackle this task.
\item We collect \ourdataset, an evaluation dataset with well-calibrated 360° video and ambisonic audio, and define a suite of semantic and spatial metrics to comprehensively assess generation quality.
\item Our method outperforms strong baselines across all metrics and in a perceptual user study on visually-guided spatial audio generation. We further apply our framework to one-shot room acoustic learning and audio-visual spatial source separation, highlighting its potential to extend to other audio-visual learning tasks in 3D.
\end{itemize}

%% file: sections/2_related.tex
\section{Related Work}
\label{sec:3d-related-o}

\paragraph{3D Scene Generation.}
Recent single-image 3D scene generation methods have coalesced into three categories: iterative, video diffusion, and panoramic, each advancing scale, coherence, and controllability in complementary ways. \emph{Iterative} methods \cite{Yu2024WonderWorldI3,Yu2023WonderJourneyGF,cai2022diffdreamer,Chung2023LucidDreamerDG,3ditscene} grow scenes by alternating diffusion-based outpainting with 3D lifting and optimization (often via Gaussian Splatting \cite{3dgs}), typically guided by geometry cues such as monocular depth and light trajectory planning with text prompts. They enable scalable outpainting and interactive refinement but can accumulate drift over long trajectories and often require post-hoc geometric cleanup. \emph{Video diffusion} methods \cite{Ren2025GEN3C3W,yu2024viewcrafter, hao2025gaussvideodreamer3dscenegeneration,feng2025wonderverseextendable3dscene,cat3d,cat4d,huang2025voyager,schneider_hoellein_2025_worldexplorer} leverage temporally coherent generators plus cached geometry such as point clouds to improve cross-view alignment, provide precise camera control, and extend from static to dynamic scenes. Their strengths include reduced flicker and controllable trajectories, while challenges include computational cost and long-range consistency when stitching extended worlds. \emph{Panoramic} methods \cite{Yang2025LayerPano3DL3,zhou2024dreamscene360,zhou2024holodreamerholistic3dpanoramic,hunyuanworld2025tencent,worldgen2025ziyangxie} first outpaint a coherent 360° panorama (typically equirectangular with upright and field-of-view constraints) and then lift it into 3D via depth alignment and Gaussian optimization, achieving full-environment coverage and seamless horizon continuity. Our approach follows this panoramic paradigm but is distinguished by jointly modeling spatial audio alongside visuals. We adopt a panoramic representation because it captures the full 360° field of view and provides a unified, scene-level coordinate frame for ambisonics rendering.

\vspace*{-1.25em}
\paragraph{Spatial Audio Generation.} 
A large body of audio generation work produces high-quality monoaural audio from either text or video inputs~\cite{mmaudio,audioldm,audioldm2,owens2016visually,zhou2017visual,chen2017,hayakawa2025mmdisco,audiogen,chen2025video}. To spatialize such audio, prior methods either explicitly model room acoustics by predicting room impulse responses (RIRs) for rendering spatial audio~\cite{avr,Jin_2025_ICCV,naf,inras,diffrir,haae}, or directly perform mono to spatial conversion conditioned on visual scene structure from videos to localize sources and synthesize spatial channels~\cite{Gao201825DVS,garg2021geometry,garg2023visually,morgado2018self,li2018scene,av-nerf}.

Most related to our work are recent methods that directly synthesizes spatial audio, either from text~\cite{Sun2024BothEW,Heydari2024ImmerseDiffusionAG} or images/videos~\cite{visage,liu2025omniaudiogeneratingspatialaudio,Dagli2024SEE2SOUNDZS}. Concurrent work Sonic4D~\cite{Xie2025Sonic4DSA} further couples spatial audio generation with 4D dynamic scenes, but it is limited to single objects, narrow views, and offline processing. In contrast, our method generates a full 3D audio-visual scene from a single image input, producing scene-consistent FOA tightly anchored to the visual context and supporting free, real-time navigation with physically grounded point, areal, and ambient sources across 360° environments, including off-screen emitters.

\vspace*{-1.25em}
\paragraph{Localizing Sounds in Visual Scenes.} Sound localization aims to identify the pixels or regions corresponding to sound sources in images or videos. Early approaches model audio-visual mutual information~\cite{fisher2001learning,hershey2000audio} or use canonical correlation analysis~\cite{kidron2005pixels}. Later deep methods exploit audio-visual correspondence with varying levels of supervision~\cite{arandjelović2018objectssound,Senocak_2018_CVPR,zhou2022audio,tian2018audio,chen2021localizing,senocak2023sound,afouras2020self,mo2022closer,hu2020discriminative}. Recently, Audio-Visual Large Language Models~\cite{zhao2023bubogpt,chowdhury2024meerkat,li2024groundinggpt} have significantly improved the generalizability of audio-visual grounding. We also localize sound sources in visual scenes; however, unlike prior work above that treats localization as the end goal, we localize and ground potential sound sources in the generated 3D panoramic scene using foundational Vision-Language Models (VLMs)~\cite{xdecoder} to enable subsequent spatial audio generation at the corresponding 3D locations.

\vspace*{-1em}
\paragraph{Audio-Visual Source Separation.} Prior work on audio-visual source separation has leveraged visual cues to guide separation of speech~\cite{gao2021VisualVoice,ephrat2018looking,chung2020facefilter}, musical instruments~\cite{Zhao2018TheSO,gao2019coseparation,gan2020music,xu2019recursive}, and general sound sources in-the-wild~\cite{gao2018objectSounds,tzinis2020into,tzinis2022audioscopev2}. Differently, our work tackles spatial audio generation for all sound sources grounded in a generated 3D visual scene, rather than separating an existing sound mixture. We further demonstrate an application to separate spatial sound sources in 3D via diffusion posterior sampling~\cite{dps}.

%% file: sections/3_task_formulation.tex
\section{The \ourtask Task}

This section reviews ambisonic spatial audio (Sec.~\ref{sec:background}), formalizes \ourtask (Sec.~\ref{sec:task_formulation}), introduces \ourdataset, our curated evaluation dataset (Sec.~\ref{sec:dataset}), and defines the evaluation metrics (Sec.~\ref{ssec:metrics}).

\subsection{Background on Ambisonics}\label{sec:background}
Ambisonics represents the sound field around \emph{a point} as a weighted sum of spherical harmonics. Let ${\bf u}(\theta, \varphi)$ denote a unit direction with azimuth $\theta\in[-\pi, \pi]$ and elevation $\varphi\in[-\pi/2, \pi/2]$, and let $a(\theta, \varphi, t)$ denote the directional sound field. Ambisonics expand this directional function on the sphere using real spherical harmonics  $Y_\ell^m(\theta,\varphi)$:
\vspace*{-0.05in}  
\begin{equation}
  a(\theta,\varphi,t)\approx\sum_{\ell=0}^{L}\sum_{m=-\ell}^{\ell}Y_\ell^m(\theta,\varphi)a_{\ell,m}(t)=\mathbf{y}_L(\theta,\varphi)^\top\mathbf{a}_L(t),
    \label{eq:ambisonics}
\vspace*{-0.05in} 
\end{equation}
where $L$ is the ambisonic order, $\mathbf{y}_L$ stacks the $(L{+}1)^2$ basis functions, and $\mathbf{a}_L(t)$ are the corresponding ambisonic coefficients (channels of sound). Given coefficients ${\bf a}_L$, a virtual microphone at direction ${\bf u}$ is obtained by $a_{\bf u}=\mathbf{y}_L({\bf u})^\top {\bf a}_{L}$. For a single point source $a_{\text{src}}(t)$ at ${\bf u}$ with distance $d$, the ambisonic coefficients at $[0,0,0]$ are:
\vspace*{-0.05in} 
\begin{equation}
    {\bf a}_L^{\text{single}}(t) =\sigma(d)a_{\text{src}}(t){\bf y}_L({\bf u}),
    \label{eq:single-source}
\vspace*{-0.05in} 
\end{equation}
where $\sigma(d)$ is the attenuation and decay.
Listener head rotation $R\in SO(3)$ enables \emph{3-DoF} rendering from a single ambisonics capture at a fixed point. Translation is not natively represented by a single-point capture. However, if ambisonics can be \emph{encoded at arbitrary listener locations} (as in our model), composing head rotations with listener positions enables full \emph{6-DoF} exploration in the sound field.

A common choice for ambisonics is first-order ambisonics (FOA; $L=1$), which has four channels corresponding to the zeroth- and first-order harmonics. FOA provides a first-order approximation of the sound field on the sphere and is widely supported in capture and playback pipelines. We use FOA in most experiments, but our rendering framework is \emph{order-agnostic} and supports arbitrary ambisonic orders $L$, trading channel count $(L{+}1)^2$ for spatial accuracy.

\subsection{Task Formulation of \textbf{\ourtask}}\label{sec:task_formulation}

Given a single input image \(I\), our objective is to develop a framework \(\mathcal{G}\) that jointly generates a visual representation \(\mathbf{V}\) and the corresponding spatial audio field \(\mathbf{A}\), forming an interactive 3D audio-visual scene:
\vspace*{-0.05in} 
\begin{equation}
    \mathcal{G}: I \rightarrow \{\mathbf{V}, \mathbf{A}\}.
\vspace*{-0.05in} 
\end{equation}
Here, \(\mathbf{V}\) encapsulates the geometric and appearance structure of the scene, while \(\mathbf{A}\) defines a spatially coherent sound field that aligns with \(\mathbf{V}\) both geometrically and semantically. For an observer at pose \(\mathbf{p}\), the scene can be rendered into an image \(\textbf{V}(\mathbf{p})\), and the corresponding spatial audio signal is given by \({\bf A}(\mathbf{p}, t)\). 

\vspace*{-0.2in} 
\paragraph{Our Instantiation.} We parameterize the visual scene ${\bf V}$ as 3D Gaussian Splats~\cite{3dgs}
(Sec.~\ref{sec:scene_generation}), and represent the auditory scene \(\mathbf{A}\) as a point-cloud-based representation that supports ambisonics rendering (Sec.~\ref{sec:foa_rendering}): given a listener pose ${\bf p}$, it synthesizes the ambisonics $ {\bf a}_{L}(t)={\bf A}({\bf p}, t)\in\mathbb{R}^{(L+1)^2}$ at arbitrary order $L$.

\subsection{The \textbf{\ourdataset} Dataset}\label{sec:dataset}

\begin{figure}[t]
\vspace*{-0.5em}
    \centering
    \includegraphics[width=1\linewidth]{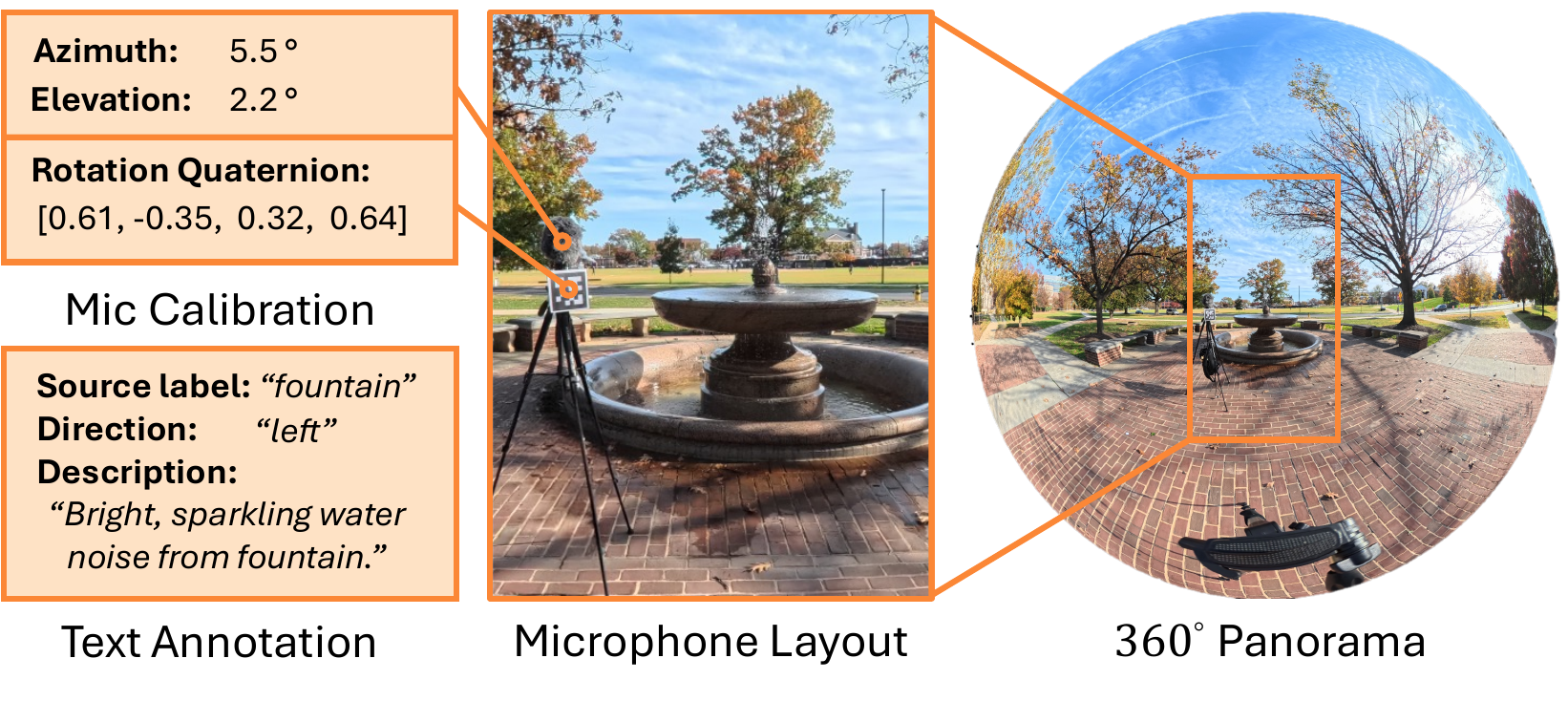}
    \vspace*{-0.3in}
    \caption{Illustration of real-world audio-visual scene data collection and curation for \ourdataset.}
    \vspace*{-1.5em}
    \label{fig:dataset-teaser}
\end{figure}

Quantitative evaluation of \ourtask requires paired 3D visual scenes with spatial audio at listener positions offset from the camera pose---conditions not met by existing datasets. Therefore, with the data collection setup shown in Fig.~\ref{fig:dataset-teaser}, we collect \ourdataset, a well-curated evaluation dataset for this task that contains 68 clips of synchronized 360° video captured by Insta360 X5 and FOA audio recorded by R{\O}DE NT-SF1 from six real-world scenes (\textit{Fountain}, \textit{Kitchen}, \textit{Pool}, \textit{Bridge}, \textit{Stream}, \textit{Siren}). See Fig.~\ref{fig:quali-real} for scene visualizations. For audible sources within each scene, we provide the semantic labels (\eg, \emph{fountain}, \emph{microwave}, \emph{bushes}), a brief description of the sound, and a coarse direction relative to the microphone, (\eg, source at \emph{left}/\emph{right}/\emph{front}/\emph{back}). These annotations enable both spatial and semantic evaluation described in Sec~\ref{ssec:metrics}. See Supp. for dataset and calibration details.

\subsection{Evaluation Metrics}\label{ssec:metrics}
\vspace*{-0.25em}
We evaluate the quality of generated spatial audio with a suite of metrics along two complementary axes: i) \emph{spatial coherence}---whether the spatial effect of sound sources in the 3D scene is accurate and realistic, and ii) \emph{semantic alignment}---whether sounds originate from the correct visually-depicted objects/regions. Below we outline each metric; see Supp. for detailed metric definitions.

\vspace*{0.025in}
\noindent\textbf{Spatial Metrics.} 
We compare the predicted spatial audio in two terms: (i) deviation in direction of arrival (DoA), and (ii) correlation of spherical energy patterns. 
After estimating DoA in azimuth $\theta$ and elevation $\varphi$, we report the absolute azimuth error $\Delta_{\text{abs}}\theta$, absolute elevation error $\Delta_{\text{abs}}\varphi$, and the geodesic angular error $\Delta_{\text{Angular}}$, following prior works~\cite{liu2025omniaudiogeneratingspatialaudio,Heydari2024ImmerseDiffusionAG}. To handle multi-source scenes where a single DoA is insufficient, we compute two saliency metrics on the spherical energy map: correlation coefficient (CC) and area under curve (AUC), following~\cite{visage}.

\vspace*{0.025in}
\noindent\textbf{Semantic Metrics.} 
To assess semantic consistency, we form directional monaural renderings by placing a virtual microphone oriented toward direction $\mathbf{u}\in\mathbb{S}^2$. Specifically, we render along four FOA-aligned principal directions: \emph{left}, \emph{right}, \emph{front}, and \emph{back}, which cover the majority of sound source placements. Using semantic labels of sound sources, we introduce two CLAP-based similarities~\cite{laionclap2023} on these directional audios: D-CLAP$_{\text{T}}$ (audio-text) between directional audio and the semantic label, and D-CLAP$_{\text{A}}$ (audio-audio) between predicted and reference directional audio. Beyond similarity, we report a directional CLAP-based R-Precision, namely D-CLAP$_{\text{R}}$, inspired by text-to-3D evaluation~\cite{jain2021dreamfields}. For annotations of the form (\text{direction}, \text{label}), we render monaural audio toward the four principal directions, rank directions by CLAP-T score for the label, and compute R-Precision (with $R=1$), \ie, the top-1 accuracy that the correct direction receives the highest similarity score.

%% file: sections/4_approach.tex
\vspace{-0.5em}
\section{Approach}
\vspace{-0.5em}
\label{sec:3d-approach-o}

\begin{figure*}
\vspace*{-1em}
    \centering
    \includegraphics[width=1\linewidth]{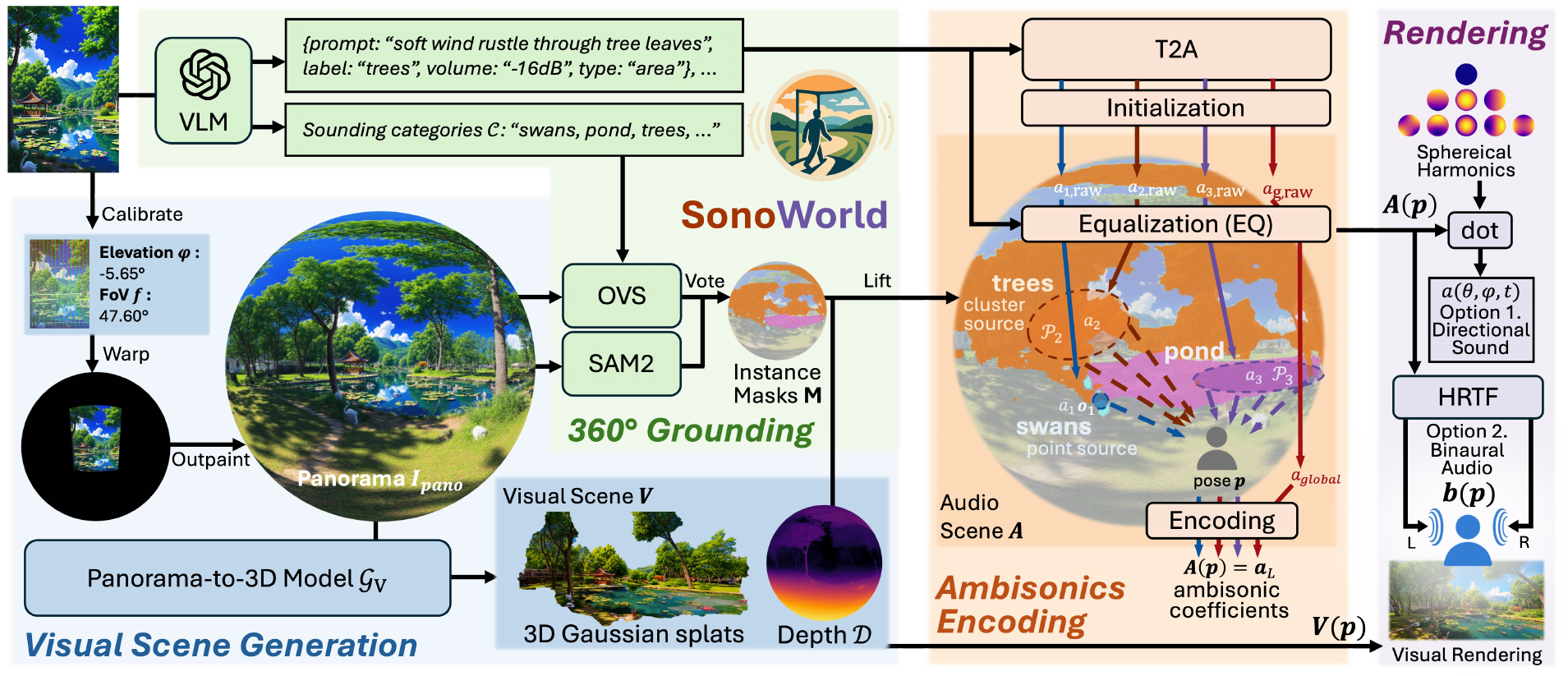}
    \vspace*{-2em}
    \caption{Given a single image $I$,  \ourframework jointly generates a 3D visual scene $\mathbf{V}$ and a semantically and geometrically aligned audio scene ${\bf A}$. It consists of: 1) \textbf{\color{vblue}{Visual Scene Generation}} (Sec. \ref{sec:scene_generation}): single-image calibration and warping followed by panorama outpainting to obtain the a full $360^\circ$ panorama image $I_{\mathrm{pano}}$, which is further lifted into a 3D Gaussian scene via a panorama-to-3D reconstruction model $\mathcal{G}_{\mathrm{\textbf{V}}}$; 2) \textbf{\color{sgreen}{360$^\circ$ Semantic Grounding}} (Sec.~\ref{sec:semantic_grounding}): a VLM extracts the categories $\mathcal{C}$ of potential sounding sources, which are used to generate panoramic instance masks $\textbf{M}$ with the help and coordination of both an open-vocabulary segmentation model (OVS) and a class-agnostic segmentation model (SAM2 \cite{ravi2024sam2}); 3) \textbf{\color{aorange}{Ambisonics Encoding}} (Sec.~\ref{sec:foa_rendering}): based on the audio prompt and equalization parameters from the VLM model, a text-to-audio (T2A) model generates per-source waveforms that are equalized, and mapped to ambisonics coefficients based on the 3D locations and its source type; and  4) \textbf{\color{ipurple} Free-Viewpoint Rendering} (Sec.~\ref{sec:inference}): the ambisonics coefficients are decoded into pose-dependent binaural audio $\textbf{b}(\textbf{p})$ using head related transfer function (HRTF) and synchronized with Gaussian rendering ${\bf V}(\textbf{p})$.}
    \label{fig:pipeline3d}
    \vspace*{-1em}
\end{figure*}

This section presents \ourframework, our \emph{training-free} pipeline to tackle the \ourtask task (illustrated in Fig.~\ref{fig:pipeline3d}): Sec.~\ref{sec:scene_generation} details how we construct a $360^\circ$ panorama and lift it into a 3D Gaussian scene $\mathbf{V}$; Sec.~\ref{sec:semantic_grounding} describes 360$^\circ$ semantic grounding to localize the sounding entities in 3D; Sec.~\ref{sec:foa_rendering} explains how we design the ambisonics encoder $\mathbf{A}$, which produces spatial sound fields that are geometrically aligned and semantically consistent with 3D visual scene $\mathbf{V}$; and  Sec.~\ref{sec:inference} demonstrates how we render binaural audio from $\mathbf{A}$ at any pose.

\subsection{Panorama-Based Visual Scene Generation}
\label{sec:scene_generation}

We adopt a panoramic representation for single-image scene generation because it naturally captures the full $360^\circ$ field of view (FoV) and unifies holistic audio-visual scene synthesis. It contains the following three steps: 1) camera calibration of the input image; 2) panoramic outpainting from the corrected image; and 3) final 3D reconstruction from the outpainted panorama, detailed below.

\vspace*{0.5em}
\noindent\textbf{Single-Image Camera Calibration.}
Prior single-image panorama outpainting methods~\cite{worldgen2025ziyangxie,hunyuanworld2025tencent,omnix} implicitly assume a level camera and place the input view along the equator of the panorama, which can cause vertical misalignment and distortion when the input image is captured with an upward or downward tilt. To address this and prepare for the outpainting step, we first use GeoCalib~\cite{veicht2024geocalib}, a single-image calibration network that jointly infers gravity direction and camera FoV through learned geometric optimization, to obtain the camera elevation and FoV as:
\vspace*{-0.05in}
\begin{equation}
(\varphi, f) = \text{Calib}(I).
\vspace*{-0.05in}
\end{equation}

\noindent\textbf{Image to Panorama Outpainting.} We then reproject the input perspective image $I$ into an equirectangular panorama using a warping operator $\mathcal{W}_G$ that rectifies the camera and performs multi-scale anti-aliased sampling based on a Gaussian pyramid. The warped image is subsequently outpainted using the panorama outpainting model $g_{\text{outpaint}}$ from WorldGen~\cite{worldgen2025ziyangxie}:
\vspace*{-0.05in}
\begin{equation}
    I_{\text{pano}} = g_{\text{outpaint}}\bigl(\mathcal{W}_G(I, \varphi, f)\bigr),
\vspace*{-0.05in}
\end{equation}
which completes the missing regions beyond the input image and produces a full $360^\circ$ panorama. Please refer to Supp. for implementation details of $\mathcal{W}_G$.

\vspace*{0.5em}
\noindent\textbf{Panorama to 3D Scene Reconstruction.} Finally, we lift the completed panorama into a 3D visual scene using a panorama-to-3D reconstruction method, denoted as $\mathbf{V} = \mathcal{G}_{\mathrm{\textbf{V}}} (I_{\text{pano}})$. 
Multiple approaches can be used for this component~\cite{Yang2025LayerPano3DL3,hunyuanworld2025tencent,worldgen2025ziyangxie,worldlabs_marble}, including World Labs' Marble model~\cite{worldlabs_marble}, which currently offers the best visual rendering quality, and the open-source model HunyuanWorld-1.0~\cite{hunyuanworld2025tencent}. Both transform the panorama into a 3D scene parameterized either by 3D Gaussian splats or a textured mesh, yielding a photorealistic 3D environment that supports real-time interactive exploration. 

\subsection{360$^\circ$ Audio-Visual Semantic Grounding}
\vspace*{-0.5em}
\label{sec:semantic_grounding}

To bridge visual and audio generation from a single image $I$, we 1) use a vision-language model (VLM) to propose sounding categories and their acoustic attributes; 2) predict instance masks on FoV images with an open-vocabulary segmenter; 3) refine them into globally consistent panoramic segmentations that align with both semantics and 3D geometry; and 4) unproject them to localize 3D sound sources in $\mathbf{V}$.

\noindent\textbf{Sounding Category Proposal.}
We first query a VLM---either the proprietary GPT-5 \cite{gpt5} or the open-source LLaVA-Next-34B \cite{liu2024llavanext}---with the input image $I$ to obtain a set of candidate sounding categories $\mathcal{C}$ and  their attributes: source-type labels (point, clustered, or ambient), text prompts for audio synthesis, and amplitude-equalization parameters. These categories then guide subsequent visual grounding.

\noindent\textbf{Open-Vocabulary Segmentation.}
Since open-vocabulary segmentation (OVS) models are trained on perspective FoV, not panoramic, images, we split the generated panorama $I_{\mathrm{pano}}$ into overlapping FoV tiles and run X-Decoder \cite{xdecoder} on each tile conditioned on $\mathcal{C}$. The resulting instance masks for each category $c \in \mathcal{C}$ are reprojected back to panoramic coordinates and grouped per category to provide semantically labeled, tile-aggregated mask predictions $\mathbf{M}_{\text{OVS},c}$.

\noindent\textbf{Panoramic Mask Refinement.} Tile-wise X-Decoder masks reprojected to the panorama are not globally consistent: limited vertical FoV and seams at tile boundaries can leave large regions (\eg, sky or ground) incomplete and introduce broken edges. In contrast, foundational segmentation models like SAM2 \cite{ravi2024sam2} can accurately segment arbitrary image formats, including equirectangular panoramas, but yield only class-agnostic regions. We therefore use SAM2 to produce panorama-wide proposals $\mathbf{M}_{\text{pano}}$, and then, for each category $c$, let the corresponding open-vocabulary predictions $\mathbf{M}_{\text{OVS},c}$ cast confidence-weighted votes on overlapping SAM2 regions based on spatial agreement. Proposals in $\mathbf{M}_{\text{pano}}$ with strong semantic support from $\mathbf{M}_{\text{OVS},c}$ are retained and, when appropriate, slightly refined to include nearby pixels endorsed by $\mathbf{M}_{\text{OVS},c}$. For each category $c$, we aggregate the retained and refined proposals into a panoramic instance set $\mathbf{M}_c$, and collect the final result as the union of all category-specific masks, $\mathbf{M} = \bigcup\limits_{c \in \mathcal{C}} \mathbf{M}_c$, preserving SAM2’s global geometric consistency while inheriting X-Decoder’s semantic precision. 

\noindent\textbf{Unprojection to 3D.} Finally, using the panoramic depth $\mathcal{D}$ rendered from the generated 3D visual scene from Sec.~\ref{sec:scene_generation}, we unproject each final panoramic instance mask $\mathcal{M}_i \in \mathbf{M}$ into 3D to obtain its spatial locations $\mathcal{P}_i = \mathrm{Lift}(\mathcal{M}_i, \mathcal{D})$. Collecting all instances yields $\mathcal{P}$ which specifies the 3D locations of all sounding objects within the scene. 

\subsection{Ambisonics Encoding}
\label{sec:foa_rendering}

Based on the VLM prediction of source type, source content, and equalization parameters, we generate an audio scene $\mathbf{A}$ and encode ambisonics at any listener pose. We use MMAudio~\cite{mmaudio} to synthesize audio from text prompts at two levels. For each grounded sound source \(i\), MMAudio generates a raw waveform \(a_{i,\mathrm{raw}}\) from a source-specific prompt. It also generates a scene-level ambient waveform \(a_{\mathrm{global}}(t)\) from a global prompt describing the overall background atmosphere. Then, we amplify the volume of each sound (equalization) via the \emph{predicted sound energy} $v$ (in dB, 0 for the loudest). The resulting audio content of source $i$ is:
\vspace*{-0.2em}
\begin{equation}
    a_i(t) = 10^{v_i/20} a_{i,\mathrm{raw}}(t).
\end{equation}

Next, we spatialize the grounded sounds \(a_i\) based on their 3D positions \(\mathcal{P}_i\) and source types predicted by the VLM. Let \(\mathcal{O}\) denote the set of grounded sound sources only. We partition \(\mathcal{O}\) into two disjoint subsets, $\mathcal{O}=\mathcal{O}_{\text{point}}\cup\mathcal{O}_{\text{cluster}}$ corresponding to point-like and clustered sources. For a listener pose \(\mathbf{p}=[\mathbf{R},\mathbf{t}]\in SE(3)\), we model distance attenuation and air absorption by $\sigma(d)=e^{-\alpha d}/d$, where $d$ is the distance from sound source to listener position ${\bf t}$. Omitting $\mathbf{p}$ and $t$ for brevity, the ambisonic coefficient is the sum of the grounded-source contributions and the global ambient term:
\vspace*{-0.25em}
\begin{equation}
\mathbf{A}
\label{eq:ambi_total}=\mathbf{A}_{\text{grounded}}+ \mathbf{A}_{\text{global}}=\mathbf{A}_{\text{point}}+
\mathbf{A}_{\text{cluster}}+\mathbf{A}_{\text{global}}
\vspace*{-0.25em}
\end{equation}

\begin{itemize}[leftmargin=*, noitemsep]
    \item \textbf{Point sources.} We approximate source $i$ by the centroid $\mathbf{o}_i$ of its point cluster $\mathcal{P}_i$, and, following Eq.~\ref{eq:single-source}, we write:
\vspace*{-0.25em}
\begin{equation}
\mathbf{A}_{\text{point}}=\sum_{i \in \mathcal{O}_{\text{point}}} a_{i, L}\sigma(\|{\bf d}_i\|)  {\bf y}_L\Bigl({\bf R}^\top\frac{{\bf d}_i}{\|{\bf d}_i\|}\Bigr),
\label{eq:point}
\vspace*{-0.25em}
\end{equation}
where ${\bf d}_i={\bf t}-{\bf o}_i$ is the relative offset vector from the listener to source $i$.

\item \textbf{Clustered sources.} For spatially extended emitters (\eg, a flowing river), we average over the point cloud $\mathcal{P}_i$ to create a diffuse field:
\vspace*{-0.25em}
\begin{equation}
\mathbf{A}_{\text{cluster}}=\sum_{i \in \mathcal{O}_{\text{cluster}}}
\frac{a_i}{|\mathcal{P}_i|}\sum_{\mathbf{o}\in \mathcal{P}_i} \sigma(\|{\bf d}\|)  {\bf y}_L\Bigl({\bf R}^\top\frac{{\bf d}}{\|{\bf d}\|}\Bigr),
\label{eq:cluster}
\vspace*{-0.25em}
\end{equation}
where ${\bf d} = {\bf t}-{\bf o}$, analogous to Eq.~\ref{eq:point}. For an area sound that surrounds the listener, the directional terms ($Y_{\ell,m}, \ell>0$) tend to cancel each other and aggregate into the omnidirectional component, making the perceived directivity less sensitive to head rotation. \item \textbf{Global ambience.} Scene-level background (\eg, wind, distant traffic) is reflected only in the omnidirectional part with no spatial variation:
\begin{equation}
\mathbf{A}_{\text{global}}= a_{\text{global}}\bigl[1,\,0,...,\,0\bigr]^\top.
\label{eq:global}
\end{equation}
\end{itemize}

The final ambisonics signal is obtained by combining grounded sources and the ambient term in Eq.~\ref{eq:ambi_total}. This encoding is fully differentiable with respect to the audio buffers \(a_i\), which makes it applicable to other 3D audio-visual learning tasks (Sec.~\ref{sec:applications}). Next, we render \(\mathbf{A}(\mathbf{p},t)\in\mathbb{R}^{(L+1)^2}\) into binaural audio for free-viewpoint rendering.

\subsection{Free-Viewpoint Rendering}\label{sec:inference}

Given a camera/microphone pose ${\bf p}$ for both image and audio synthesis, we render synchronized visuals in images and  encode spatial sound in ambisonics. The visual frame $\textbf{V}(\textbf{p})$ is obtained by standard 3D Gaussian Splatting rendering. The ambisonics signals $\mathbf{a}_L(t)$ are  decoded to binaural output via an Head Related Transfer Function (HRTF)-based decoder. Let $h_{\ell,m}^{\text{left}}$ and $h_{\ell,m}^{\text{right}}$ denote the left/right HRIRs for ambisonics channel $m,\ell$. The final binaural waveform $\mathbf{b}(${\bf p}$)=[b_{\text{left}}, b_{\text{right}}]^\top$ is computed as: 
\vspace*{-0.25em}
\begin{equation}
\begin{bmatrix} b_{\text{left}}\\ b_{\text{right}} \end{bmatrix}(t)
= \sum_{\ell=0}^L \sum_{m=-\ell}^\ell
\begin{bmatrix}
h_{\ell,m}^{\text{left}} * {a}_{\ell, m} \\
h_{\ell,m}^{\text{right}} * {a}_{\ell, m}
\end{bmatrix}(t),
\label{eq:binaural}
\vspace*{-0.25em}
\end{equation}
where $*$ denotes convolution. Updating the camera and microphone pose $\mathbf{p}$ enables interactive navigation with view-consistent, spatialized audio. See Supp. for details.

%% file: sections/5_results.tex
\section{Experiments}
\label{sec:3d-results-o}

\subsection{Experiment Settings}
\paragraph{Dataset.}
Since no existing dataset provides ground-truth spatial audio recorded at listener positions distinct from camera viewpoints, we rely on our \ourdataset~dataset (Sec.~\ref{sec:dataset}) for quantitative evaluation. For qualitative evaluation, user studies, and demos, we additionally use online photographs and diffusion-generated images, spanning a wide range of audio-visual events (\eg, plane landing, riverside market, volcano eruption) that are often difficult to capture in real-world recordings. Together, these real-world and synthetic scenes provide a complementary and comprehensive evaluation of generation quality.

\vspace*{-1.5em}
\paragraph{Baselines.} No existing approach directly addresses the \ourtask task we propose. We compare with prior methods that generate spatial audio from visual input (SEE-2-SOUND~\cite{Dagli2024SEE2SOUNDZS}, ViSAGe~\cite{visage}, OmniAudio~\cite{liu2025omniaudiogeneratingspatialaudio}), as well as MMAudio~\cite{mmaudio}, a representative monaural audio generation method. For comparison, we adapt each method to our setting by feeding it the 3D visual scene rendered by \ourframework and using it to generate spatial audio, while keeping their archetectures unchanged. Below, we describe the specific adaptation for each baseline.

\begin{itemize}

    \item \textbf{MMAudio~\cite{mmaudio}}: It generates monaural audio conditioned on a FoV video, which we render from our reconstructed scene. We further guide it with a text prompt composed of all detected sounding categories and directly pan the synthesized audio sources using their ground-truth locations to obtain spatial audio, both to the baseline's advantage.

    \vspace*{0.25em} 
    
    \item \textbf{SEE-2-SOUND~\cite{Dagli2024SEE2SOUNDZS}}: It is designed for FoV image inputs and lifts audio anchors into 3D using per-region depth estimates. We render FoV images from our reconstructed 3D scene to guide its ambisonics generation.
    
    \vspace*{0.1em}
    \item \textbf{ViSAGe~\cite{visage}}: generates ambisonics from a FoV video, which we render from our reconstructed scene. 

    \vspace*{0.1em}

    \item \textbf{OmniAudio~\cite{liu2025omniaudiogeneratingspatialaudio}}: generates ambisonics from a panorama video, which we render from our reconstructed scene.
    
    \vspace*{-0.5em}

\end{itemize} 

\subsection{Quantative Results}

\paragraph{Quantitative Evaluation on \textbf{\ourdataset}.}

\begin{figure}[t]
    \vspace{-0.5em}
    \centering
    \includegraphics[width=.98\linewidth]{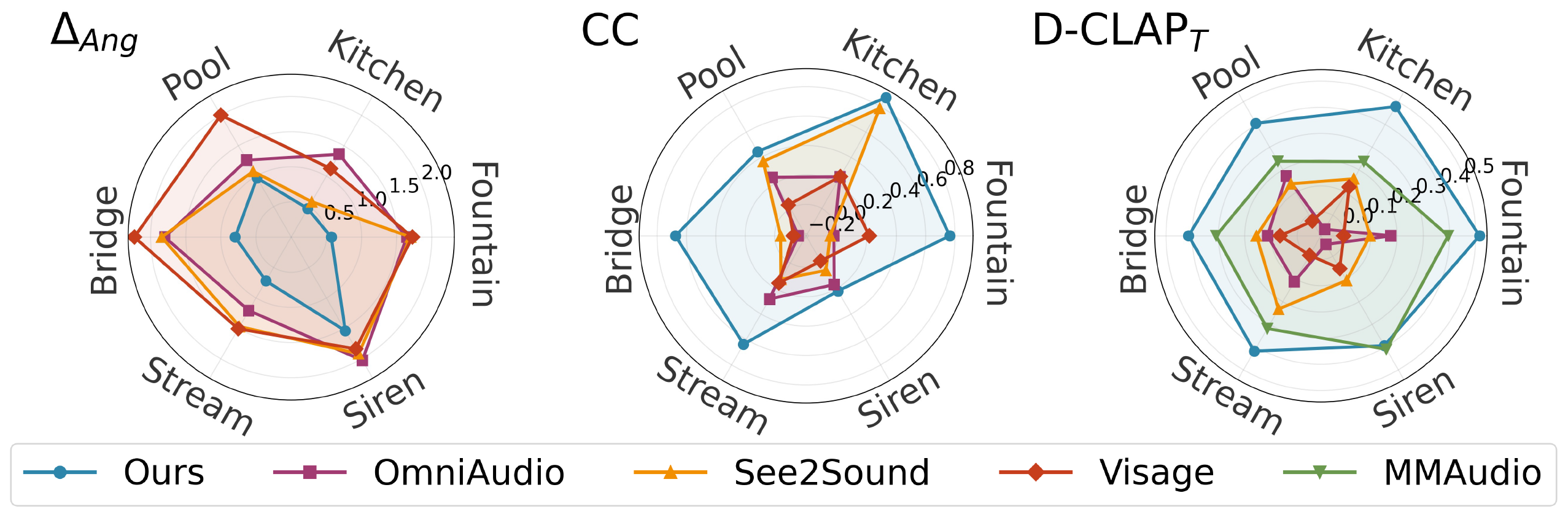}
    \vspace*{-1.0em}
    \caption{\textbf{Per-scene results on our \ourdataset dataset.} Results on representative metrics
    show that our method consistently outperforms baselines on all scenes.}
    \vspace*{-1.5em}
    \label{fig:per_scene_breakdown}
\end{figure}

Table~\ref{tab:quantitative} reports results on both spatial and semantic metrics introduced in Sec.~\ref{ssec:metrics} on \ourdataset. We report two versions of our results: \textbf{Ours (Open-source)}, a fully reproducible open-source version, where we use HunyuanWorld-1.0~\cite{hunyuanworld2025tencent}
for panorama-to-3D scene reconstruction and LLaVA-Next-34B \cite{liu2024llavanext} for sound-source proposal; and \textbf{Ours (Proprietary)}, which instead uses Marble \cite{worldlabs_marble} and GPT-5 \cite{gpt5} for the best performance. Our approach consistently outperforms all baselines across every metric: compared to all spatial audio generation baselines, we reduce DOA error by 47\% and improve CC and AUC by  \textgreater 239\% and 34\%. On semantic metrics, our method achieves higher D-CLAP scores, improving them by more than 39\% even relative to the state-of-the-art video-to-audio baseline MMAudio, and by more than 117\% over spatial-audio baselines. Note that all baselines benefit from outputs produced by proprietary models, yet our method still significantly outperforms them even in the open-source setting. 
In addition, we estimate the uncertainty via 95\% confidence intervals computed as $1.96\times\mathrm{SEM}$
(standard error of the mean): 
$\Delta_{\text{Angular}}=0.728\pm0.100$, CC$=0.658\pm0.063$, and $\text{D-CLAP}_{\text{T}}=0.457\pm0.014$, indicating stable gains. Fig.~\ref{fig:per_scene_breakdown} shows per-scene results. See Supp. for ablation studies.

\vspace*{-1.2em}
\paragraph{User Study.} 
We conduct a user study with 50 participants across 12 scenes (6 real scenes from SonoScene360 and 6 synthetic scenes), comparing our method with MMAudio \cite{mmaudio} and Omniaudio \cite{liu2025omniaudiogeneratingspatialaudio}. 
For each scene, we render a fixed-trajectory video from our generated visual scene and pair it with each method's spatial audio. Participants perform three pairwise comparisons (Ours vs.\ MMAudio, Ours vs.\ Omniaudio, and Omniaudio vs.\ MMAudio), selecting the preferred video for each pair based on both spatial coherence and audio-visual semantic alignment. Note that the visuals are identical across methods in the user study; only the spatial audio differ. 
We report pairwise preference rates averaged across scenes. 
Across both real and synthetic scenes, our method achieves the highest human preference, in line with quantitative evaluation results above. 

\begin{figure}
\vspace*{-0.5em}
    \centering
    \includegraphics[width=0.9\linewidth]{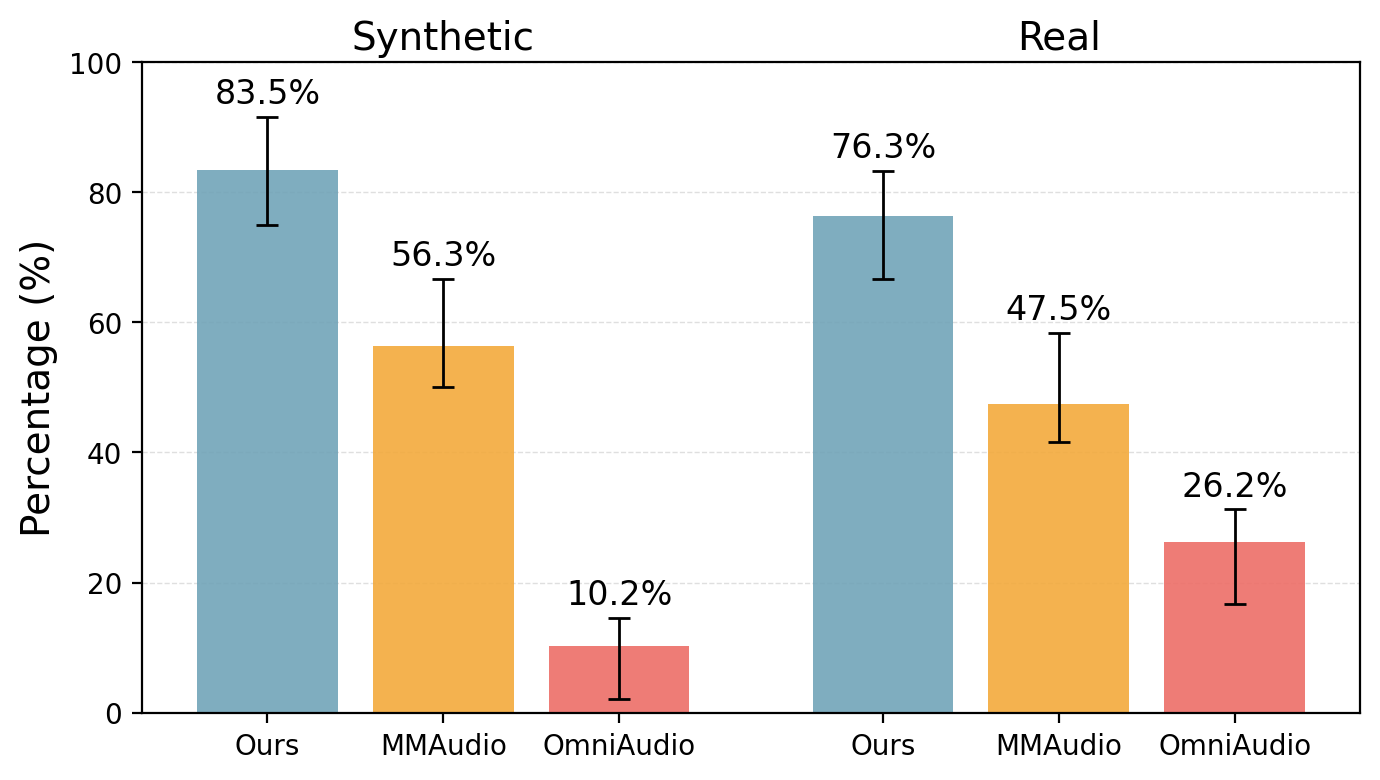}
    \vspace*{-0.5em}
    \caption{\textbf{User Study of Spatial Audio Quality.} Preference rates across \emph{synthetic} and \emph{real} scenes for our method, MMAudio~\cite{mmaudio}, and OmniAudio~\cite{liu2025omniaudiogeneratingspatialaudio}. Each bar shows the average per-user, per-scene preference (\%); error bars indicate the interquartile range (25\textsuperscript{th}--75\textsuperscript{th} percentile) across 50 participants.
    }
    \label{fig:user_study}
    \vspace*{-1.8em}
\end{figure}

\subsection{Qualitative Results}

\paragraph{Free-Viewpoint Audio-Visual Exploration.} 
Our framework enables interactive free-viewpoint navigation in the generated 3D audio-visual scenes, with an audio callback under $1\,\mathrm{ms}$ on an Apple M3 Pro laptop for the \emph{Fountain} scene, well within the $5.3\,\mathrm{ms}$ latency of a 256-sample buffer at 48kHz, easily meeting real-time constraints. We further implement a public browser-based viewer in which visuals are rendered with Three.js and spatial audio is synthesized via the WebAudio API, running entirely on a standard laptop CPU.
Please check our project website for details of our interactive interface and video demos across diverse scenes.

\input{assets/tables/table-2-numerical-results-main}

\input{assets/tables/table-3-room-acoustic}

\begin{figure*}
    \centering
    \includegraphics[width=\linewidth]{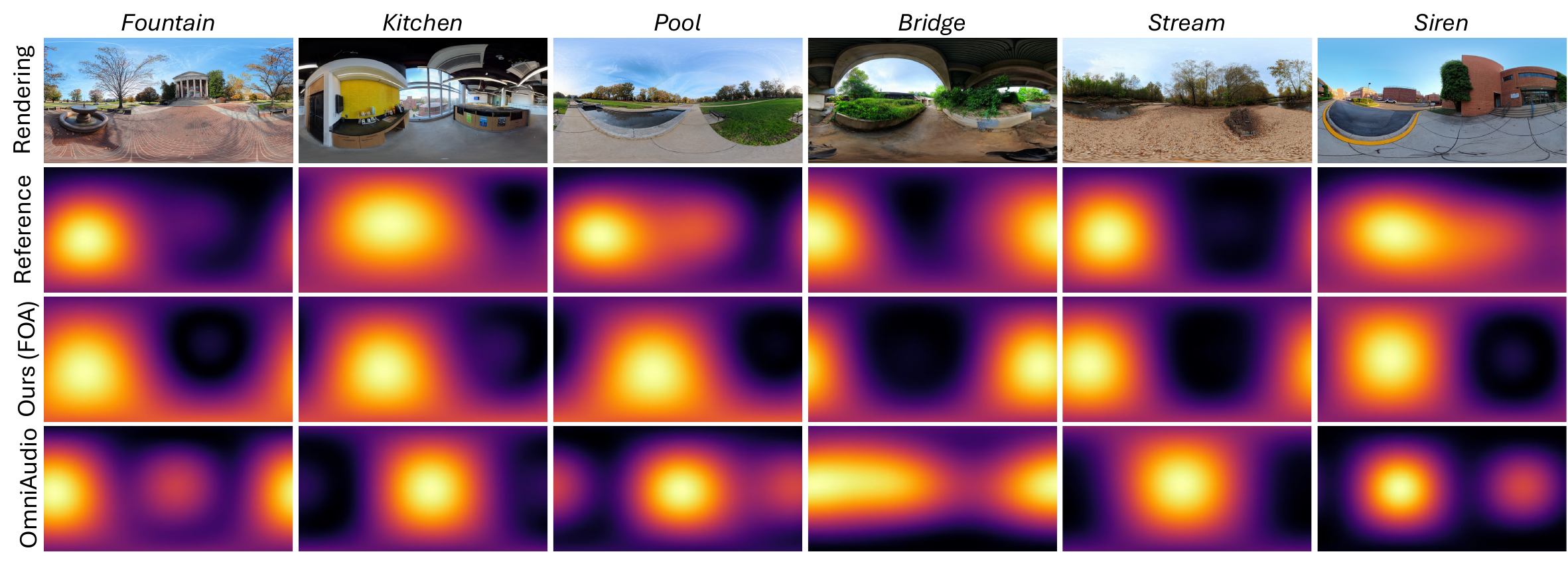}
    \vspace*{-2em}
    \caption{
    \textbf{Ambisonics energy map visualization on real-world scenes.} Top row: panoramic renderings from our method at novel views (microphone poses) on \ourdataset. Second row: ground-truth FOA energy maps. Remaining rows: FOA predictions from our method and OmniAudio~\cite{liu2025omniaudiogeneratingspatialaudio}. Our maps most closely follow the ground-truth reference spatial patterns, while the baselines produce oversmoothed or misaligned energy. For more qualitative baseline comparisons, please refer to Supp. 
    }
    \label{fig:quali-real}
    \vspace*{-1.0em}
\end{figure*}

\vspace*{-1.5em}
\paragraph{Ambisonics Energy Map Visualization.}

We visualize per-direction ambisonics energy maps in Fig.~\ref{fig:quali-real} for scenes in \ourdataset. The columns correspond to different scenes and rows to ambisonics predictions from different methods, with brighter regions indicating higher sound energy. On these real scenes, our FOA maps best match the ground truth, with high-energy lobes aligned to the correct directions and spatial extents of sounding objects, while OmniAudio yields oversmoothed or weakly correlated patterns. Please refer to Supp. for qualitative results on synthetic data with camera movements. The \textit{Siren} scene in Fig.~\ref{fig:quali-real} shows a typical failure case, where the source (police car) is moving; since our method takes a static image as input, it has no cues about source motion.

\subsection{Extensions and Applications}\label{sec:applications}

A key property of our framework is that the entire spatial audio rendering pipeline is differentiable, connecting source audio and scene parameters to the resulting spatial audio field via analytic gradients. This enables direct optimization of scene- and source-level variables from audio observations, opening up a range of 3D audio-visual learning tasks beyond scene generation. We highlight two such extensions---\emph{one-shot room acoustic learning} and \emph{audio-visual spatial source separation}---as examples of broader 3D audio-visual applications  enabled by our formulation. Please see Supp. for the detailed setup of these two tasks.

\vspace*{-1em}
\paragraph{One-Shot Room Acoustic Learning.}
Given a synthesized 3D scene with fixed geometry and visually localized sources, we optimize the parameters of our differentiable renderer such that the predicted ambisonics match a target FOA recording at one microphone pose. Tab.~\ref{tab:acousticlearning} shows that our method substantially outperforms neural acoustic field baselines NAF~\cite{naf}, INRAS~\cite{inras}, and AV-NeRF~\cite{av-nerf} on all metrics, indicating that our differentiable spatial audio renderer can serve as an effective and data-efficient surrogate for room acoustics, even in the challenging one-shot setting.

\vspace*{-1.5em}
\paragraph{Audio-Visual Spatial Source Separation.}

We further treat our renderer as a differentiable spatialization module for audio-visual source separation. Given a mixture spatial recording and visually localized sources in 3D, we estimate per-source waveforms whose rendered sum matches the observed mixture, encouraging each source to explain energy near its corresponding visual region. This provides a spatially grounded separation objective built directly in 3D. We show that we can perform source separation in 3D on FOA recording based on the semantic understanding of the visual scene. See Supp. for demonstration.

%% file: assets/tables/table-2-numerical-results-main.tex
\begin{table*}[t]
\vspace*{-2em}
\begin{small}
    \centering
    \begin{tabular}{lcccccccc}
    \toprule
    & \multicolumn{5}{c}{Spatial Metrics} & \multicolumn{3}{c}{Semantic Metrics}\\
    \cmidrule(lr){2-6}  \cmidrule(lr){7-9}
Method & $\Delta_{\text{abs}}\theta$ $\downarrow$ & $\Delta_{\text{abs}}\varphi$ $\downarrow$ & $\Delta_{\text{Anglular}}$ $\downarrow$ & CC $\uparrow$ & AUC $\uparrow$ & D-CLAP$_{\text{R}}$ $\uparrow$ & D-CLAP$_{\text{A}}$ $\uparrow$ & D-CLAP$_{\text{T}}$ $\uparrow$\\
    \midrule
MMAudio~\cite{mmaudio}  & --- & --- & --- & --- & --- & 33.8\% & 0.345 & 0.322 \\
SEE-2-SOUND~\cite{Dagli2024SEE2SOUNDZS}& 1.354 & 0.254 & 1.397 & 0.194 & 0.603 & 22.1\% & 0.221 & 0.156\\
ViSAGe~\cite{visage}    & 1.598 & 0.426 &1.649 & 0.142 & 0.624 & 22.1\% & 0.088 & 0.019  \\
OmniAudio~\cite{liu2025omniaudiogeneratingspatialaudio}    & 1.508 & 0.317 & 1.449 & 0.148 & 0.588 & 39.7\% & 0.101 & 0.104 \\
\midrule
Ours (Open-source) & 0.976 & 0.251 & 0.975 & 0.491 & 0.753 & 52.9\% & 0.464 & 0.413\\
Ours (Proprietary) & \textbf{0.672} & \textbf{0.216} & \textbf{0.728} & \textbf{0.658} & \textbf{0.838} & \textbf{67.6\%} & \textbf{0.480} & \textbf{0.457} \\
    \bottomrule
    \end{tabular}
    \vspace*{-0.5em}
    \caption{\textbf{Quantitative comparison on \ourdataset.} We report two versions of our method: \textbf{Ours (Open-source)}, a fully reproducible version using only open-source models, and \textbf{Ours (Proprietary)}, which leverages proprietary models for the best performance. }
    \label{tab:quantitative}
    \end{small}
    \vspace*{-0.5em}
\end{table*}

%% file: assets/tables/table-3-room-acoustic.tex
\begin{table}[t]
\small
    \centering
    \begin{tabular}{lccc}
    \toprule
       Method  & $\Delta_{\text{Angular}}$ $\downarrow$ & MAG $\downarrow$ & ENV $\downarrow$\\
     \midrule
        NAF~\cite{naf} & 1.76 & 3.96 & 3.60 \\
        INRAS~\cite{inras} & 1.64 & 5.06 & 5.97 \\ 
        AV-NeRF~\cite{av-nerf} & 1.58 & 4.58 & 1.89  \\
        Ours & \textbf{0.22} & \textbf{3.46} & \textbf{1.22} \\
     \bottomrule
    \end{tabular}
    \caption{Results on one-shot room acoustic learning.}
    \label{tab:acousticlearning}
    \vspace*{-2em}
\end{table}

%% file: sections/6_conclusion.tex
\section{Conclusion}
We presented \ourtask---generating a 3D audio-visual scene from a single image---and \ourframework, the first training-free framework for this task. Our method produces explorable scenes that support free-viewpoint audio-visual rendering:  photorealistic novel views paired with perceptually realistic ambisonics aligned with scene geometry and sound source semantics. Quantitative evaluation on spatial audio generation and a user study both confirm the effectiveness of our proposed method. Beyond the core task, we showed preliminary extensions of our differentiable design to one-shot acoustic learning and audio-visual spatial source separation. As future work, we aim to extend our framework to more challenging 3D audio-visual learning tasks and dynamic scenes with moving sources.

\vspace*{0.5em}
\noindent \small
{\bf Acknowledgments:} This project was supported in part by a gift from Dolby Laboratories and Barry Mersky and Capital One Endowed Professorships. We thank the anonymous reviewers for their useful feedback that helped improve this manuscript.

%% file: supp.tex
\maketitlesupplementary
\startcontents[supp]

{
    \hypersetup{linkcolor=black}
    \setcounter{tocdepth}{2}
    \small
    \setlength{\baselineskip}{1.2\baselineskip}

    \titlecontents{section}
      [2.0em]
      {\bfseries}
      {\contentslabel{1.8em}}
      {}
      {\hfill\contentspage}
    
    \titlecontents{subsection}
      [3.6em]
      {}
      {\contentslabel{2.2em}}
      {}
      {\titlerule*[1.0pc]{.}\contentspage}

    \section*{Contents}
    \printcontents[supp]{}{1}{}
}

\section{Supplementary Video}

The supplementary video provides a visual and auditory overview of \ourframework and its applications. It is organized into the following parts:

\begin{enumerate}
    \item \textbf{Task and pipeline recap.}
    We first briefly recap the problem setting of generating a 3D audio-visual scene from a single image, and summarize our overall pipeline, including visual scene generation, panorama grounding, spatial audio encoding, and free-view rendering.

    \item \textbf{Interactive web demo.}
    We then show the layout of our interactive web demo, where users can freely move the listener in 3D and rotate the head while listening to spatialized audio in real time. This section includes five screen recordings of live interactions to highlight responsiveness and stability.

    \item \textbf{Baseline qualitative comparison.}
    Next, we present qualitative comparisons with OmniAudio~\cite{liu2025omniaudiogeneratingspatialaudio} and MMAudio~\cite{mmaudio} on the same scenes. For each scene, we fix the camera trajectory and compare how different methods generate spatial audio.

    \item \textbf{Long-trajectory visualizations.}
    We show longer camera trajectories rendered with our method with rendered binaural audio together with visualizations of the FOA direction of arrival (DoA) as the listener moves through the 3D scene.

    \item \textbf{Extension: one-shot room acoustic learning.}
    We demonstrate the one-shot room acoustic learning setup, where our differentiable renderer is fit to a single source--listener FOA recording. The video shows how the learned room response generalizes to new listener positions and to new source audio played through the same scene.

    \item \textbf{Extension: audio-visual spatial source separation.}
    Finally, we showcase audio-visual spatial source separation on a YouTube 360$^\circ$ video with FOA audio. Given the visual layout and the recorded spatial mixture, our method separates the spatial audio into monaural tracks, for example isolating a violin and a beatboxer from the same 360$^\circ$ performance.
\end{enumerate}

\section{Ablation Studies}
\input{assets/tables/supp-ablation-}

Table~\ref{tab:ablation} evaluates the contribution of key components in our framework (\emph{Ours (Proprietary)} in main). First, modeling sources as finite regions rather than point emitters is important for spatial accuracy: \textbf{Ours (Point)} degrades substantially on all spatial metrics, indicating that extended source support is necessary to capture realistic directional spread. Second, the equalization module improves both spatial consistency and semantic alignment: removing it (\textbf{Ours (w/o EQ)}) yields noticeably worse $\Delta_{\text{Angular}}$, CC, AUC, and D-CLAP$_{\text{T}}$, showing that explicit per-source amplitude correction is important for matching the rendered scene geometry with the generated audio content. Third, our SAM2-based mask voting strategy is critical for robust visual grounding. Both \textbf{Ours (No merge)}, which treats tile masks independently, and \textbf{Ours (All merge)}, which merges all masks of the same category, underperform the full system by a clear margin, confirming that our voting-and-merging design yields more reliable source extents and locations (See Figure~\ref{fig:vo} for qualitative comparisons). Finally, we test robustness to imperfect geometry and segmentation by perturbing mask boundaries and depth maps. Both \textbf{Ours (Boundary perturb)} and \textbf{Ours (Depth perturb)} incur only minor performance drops relative to the full model, indicating that our method is stable under realistic noise in visual grounding and 3D reconstruction. Overall, the full system achieves the best balance of spatial and semantic quality.

\section{More Qualitative Results}

\subsection{Synthetic and Real-World Scenes}

In Fig.~\ref{fig:quali_synth}, we visualize \ourframework on two synthetic scenes generated by a text-to-image diffusion model. From a single input image (top row), our method reconstructs a 3D Gaussian scene and predicts a spatial audio field that can be queried along arbitrary camera trajectories. We show several rendered views along the trajectory together with the corresponding first order ambisonic (FOA) and second order ambisonic (SOA) spherical energy maps. In the \emph{Garden} scene, the energy clearly concentrates around the waterfall and two streams as the camera moves, while in the \emph{Riverside Market} scene the energy follows the visually salient market area, illustrating that our model can localize multiple extended sources in synthetic environments.

\begin{figure}[t!]
    \centering
    \includegraphics[width=\linewidth]{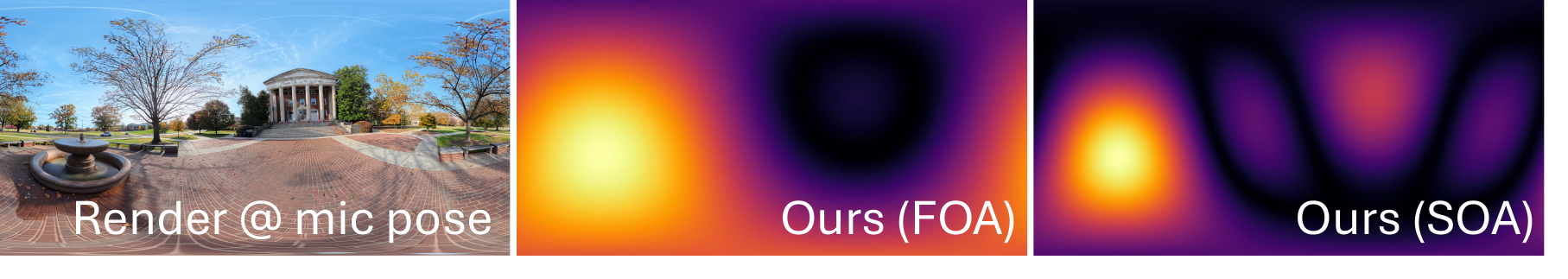}
    \vspace{-1.5em}
    \caption{The sound energy by second order ambisonics (SOA) in real scenes.}
    \label{fig:real_soa}
    \vspace{-1.5em}
\end{figure}

\begin{figure*}
    \centering
    \includegraphics[width=\linewidth]{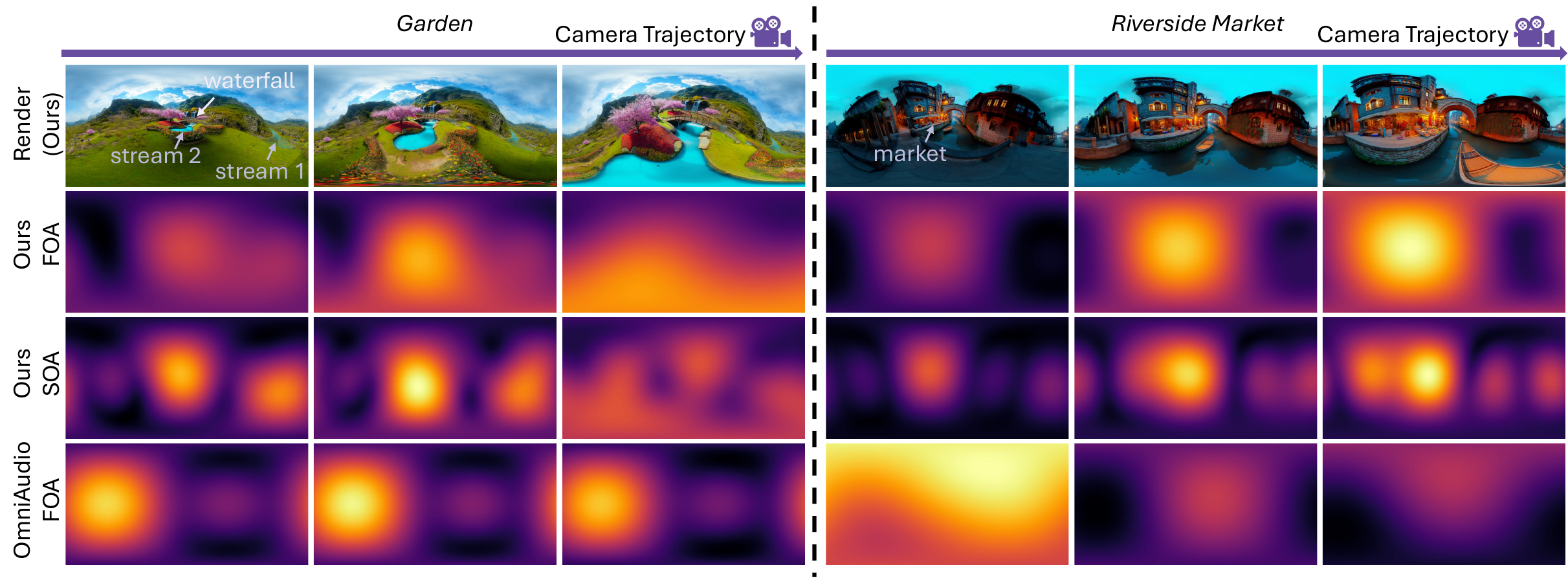}
    \caption{\textbf{Qualitative results on synthetic scenes.}
    From a single diffusion-generated image, \ourframework reconstructs a 3D Gaussian scene and predicts a spatial audio field that supports free-viewpoint exploration. We show two synthetic scenes (\emph{Garden} and \emph{Riverside Market}), sample views along a camera trajectory (top row), and the corresponding FOA / SOA spherical energy maps for our method and OmniAudio. The energy smoothly tracks visually grounded sources such as the waterfall, streams, and market stalls.}
    \label{fig:quali_synth}
\end{figure*}

\begin{figure*}
    \centering
    \includegraphics[width=\linewidth]{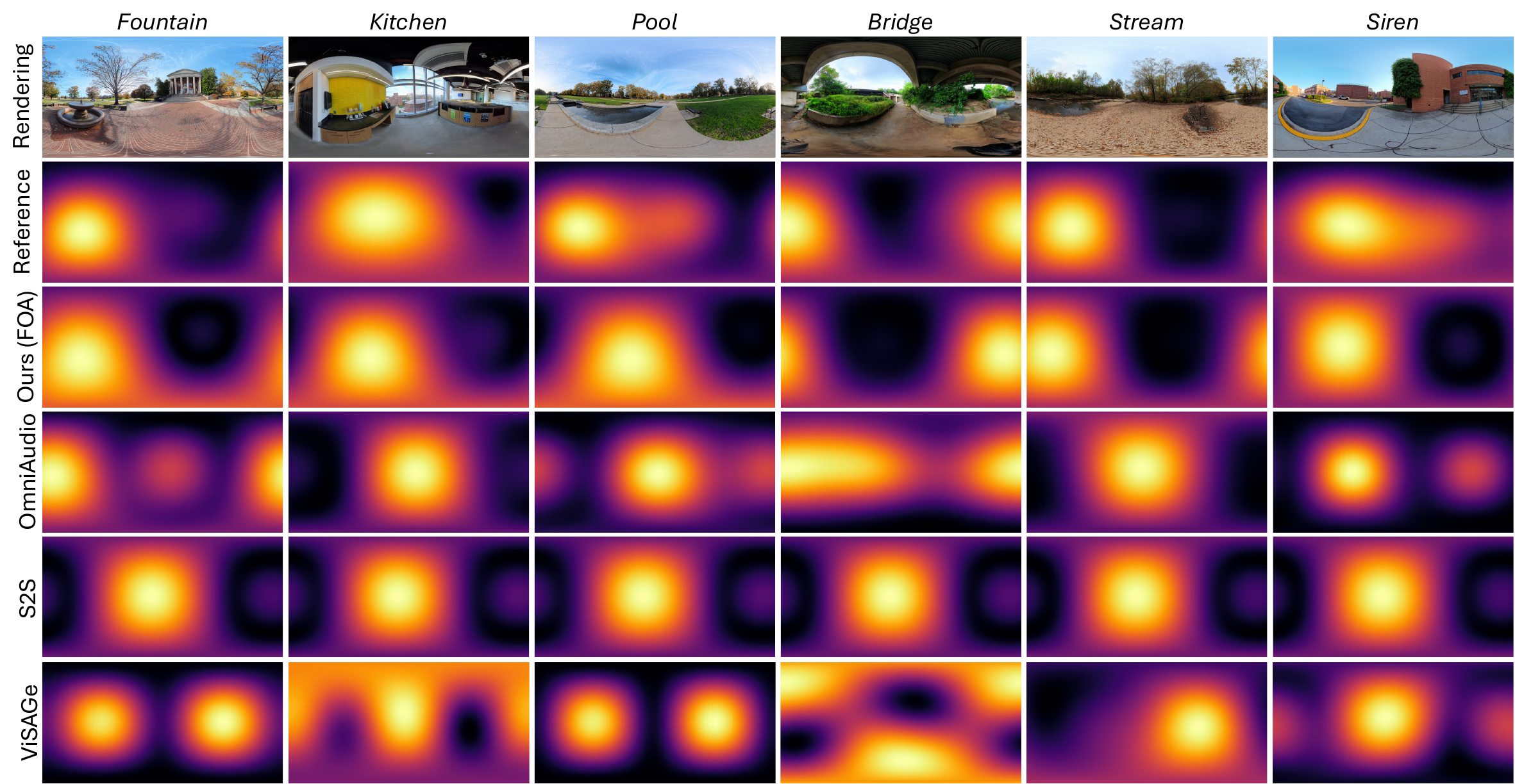}
    \caption{\textbf{Qualitative comparison on real scenes.}
    Additional qualitative results on six real environments from \ourdataset. For each scene, we show the rendered panorama from our 3D Gaussian reconstruction, the reference spherical energy map derived from the recorded FOA (second row), and the predicted energy maps from our method (FOA), OmniAudio, SEE-2-SOUND (S2S), and ViSAGe. Our method produces spatial patterns that best align with the reference both in azimuth and elevation and in the spatial extent of the energy, especially around extended water and ambient sources. Please refer to the \emph{supplementary video} for audio.}
    \label{fig:quali_real}
\end{figure*}

Figure~\ref{fig:quali_real} provides additional qualitative comparisons on real scenes from \ourdataset. For each environment, we show a rendered panorama of the reconstructed 3D scene, the reference spherical energy map computed from the recorded FOA, and the energy maps from our method and three adapted baselines (OmniAudio~\cite{liu2025omniaudiogeneratingspatialaudio}, SEE-2-SOUND (S2S)~\cite{Dagli2024SEE2SOUNDZS}, and ViSAGe~\cite{visage}). Our predictions most closely match the reference both in the dominant direction of arrival and in the spread of energy around extended sources (e.g., water in \emph{Fountain} and \emph{Stream}, room ambience in \emph{Kitchen}), while some baselines either over-smooth the field or collapse it to overly concentrated blobs. Please refer to the \emph{supplementary video} for the accompanying audio. In Fig.~\ref{fig:real_soa}, we show the 
in real scenes, SOA also yields sharper, more concentrated energy maps around the sounding objects.

\subsection{3D Scene Backbones}

\begin{figure*}[t]
    \centering
    \includegraphics[width=.8\linewidth]{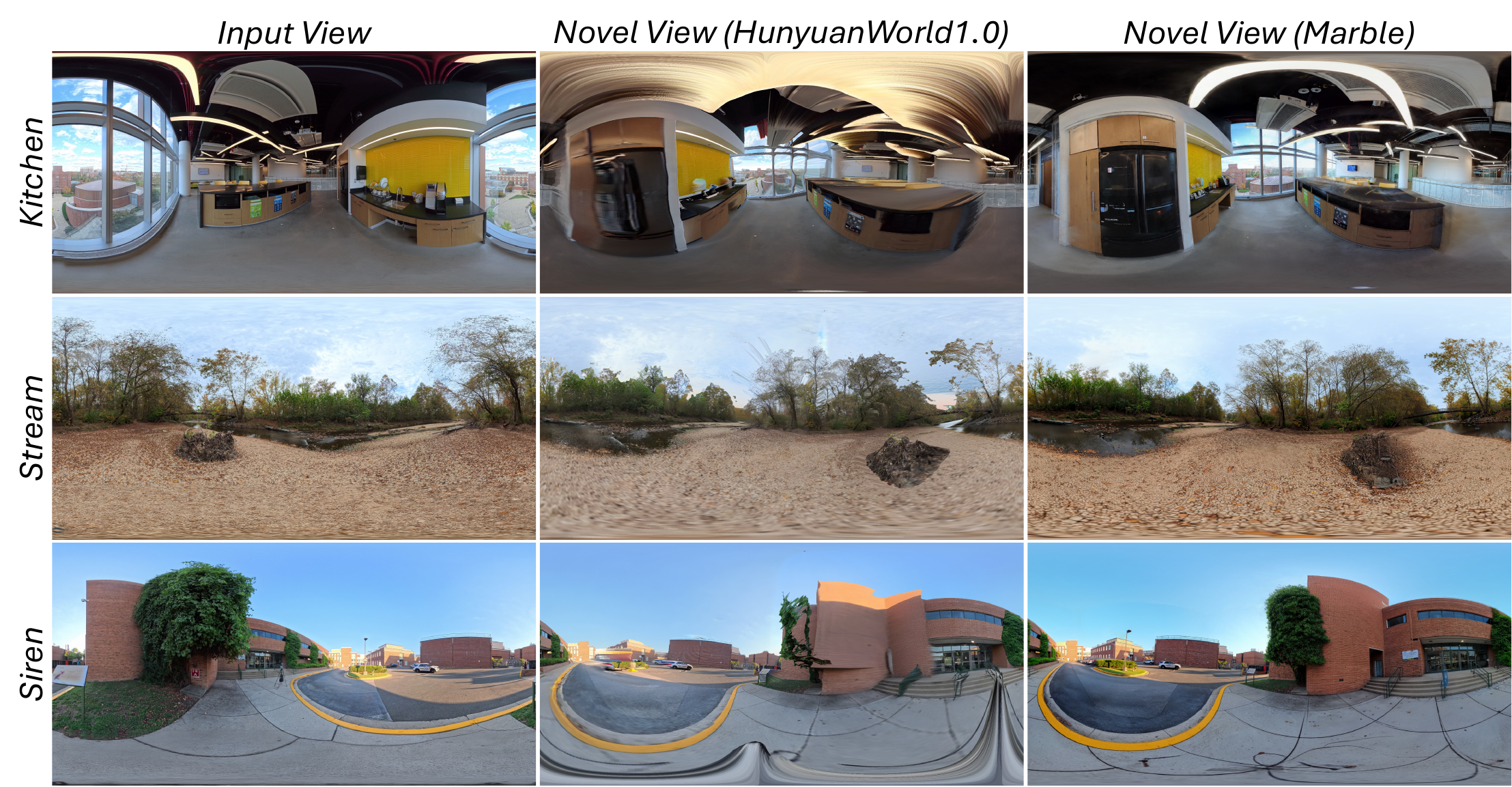}
    \caption{\textbf{Comparison of 3D scene backbones.}
    Given an input 360$^\circ$ panorama (left column), we compare novel views generated by HunyuanWorld~1.0 and Marble for several real scenes (\emph{Kitchen}, \emph{Stream}, \emph{Siren}). HunyuanWorld produces globally coherent but sometimes distorted geometry (e.g., warped floors and façades), whereas Marble yields sharper, more faithful reconstructions that preserve local spatial layout, which is crucial for accurate spatial audio anchoring and free-viewpoint navigation.}
    \label{fig:3dcomp}
\end{figure*}

We next analyze the impact of different 3D scene backbones on the visual quality of the reconstructed environments. Figure~\ref{fig:3dcomp} compares HunyuanWorld~1.0~\cite{hunyuanworld2025tencent} (with mesh output) and Marble~\cite{worldlabs_marble} (with mesh or 3DGS, the renderings are from 3DGS) when conditioned on the same input panorama. For each real scene, we show the input view (left) and one novel view generated by HunyuanWorld and Marble.

HunyuanWorld often produces visually rich global structure but can introduce noticeable distortions and oversmoothing in nearby geometry, which is undesirable for precise audio anchoring. Marble, in contrast, yields sharper details and more faithful local geometry with fewer distortions around important objects (e.g., kitchen counters, river banks, and buildings), while still enabling efficient real-time rendering. These observations support our choice of Marble as the default backbone in the main experiments.

\subsection{Panoramic Instance Merging}

\begin{figure*}[t]
    \centering
    \includegraphics[width=.8\linewidth]{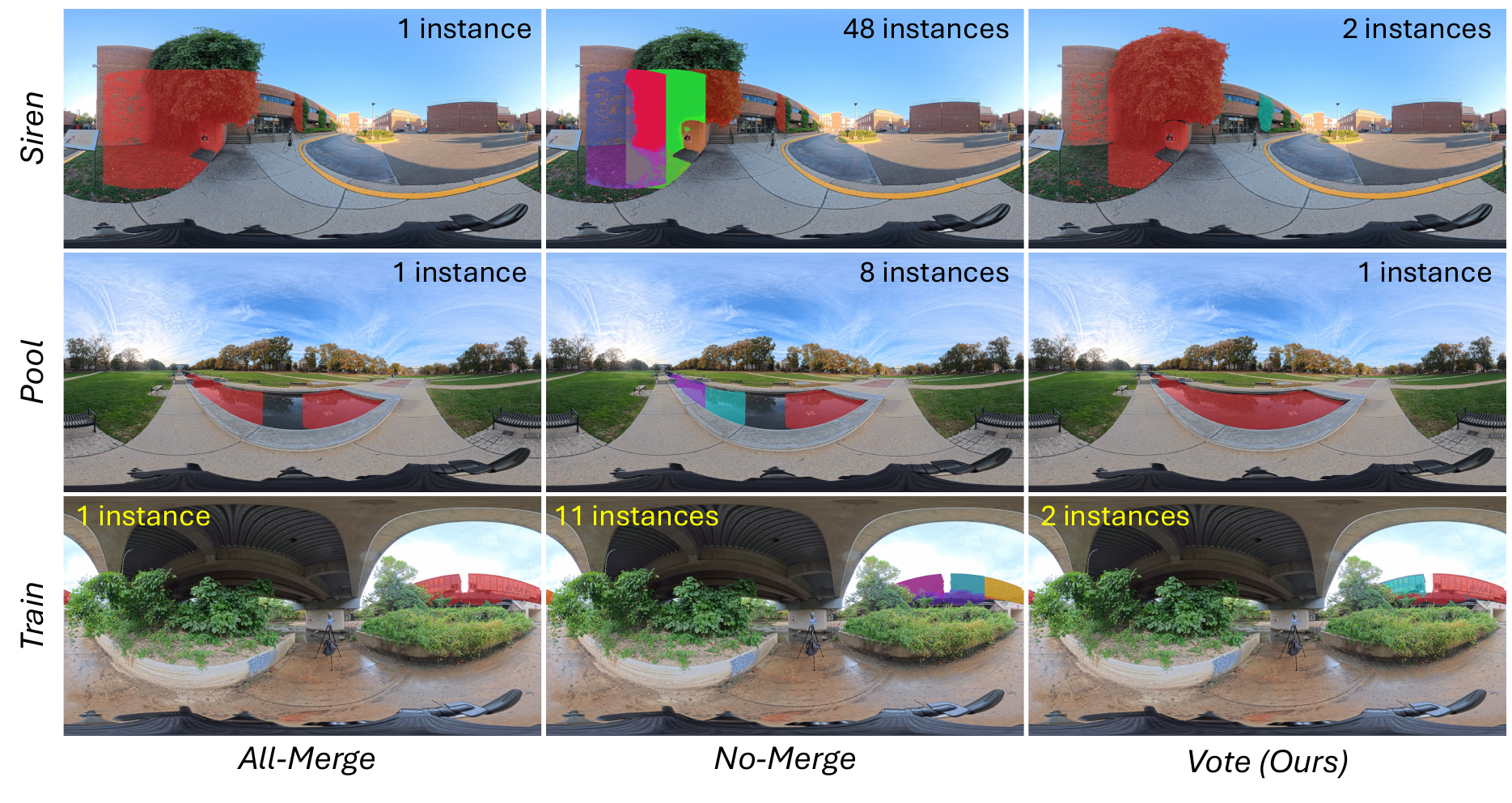}
    \caption{\textbf{Panoramic instance merging strategies.}
    We compare three strategies for grouping class-agnostic region proposals into semantic instances in \ourdataset: \emph{All-Merge} (left), \emph{No-Merge} (middle), and our voting-based strategy \emph{Vote (Ours)} (right). For each scene, we overlay the resulting instances on the panorama and indicate their counts. All-Merge collapses large structures into a single instance, losing spatial detail; No-Merge over-fragments the scene into many small pieces. Our voting scheme strikes a balance, producing a small number of coherent instances that align with visually meaningful sound sources.}
    \label{fig:vo}
\end{figure*}

Our panoramic grounding pipeline combines class-agnostic proposals (from SAM2-style segmentation) with open-vocabulary semantic masks to derive sound source instances. A key design choice is how to merge or split the underlying regions into semantic instances. Figure~\ref{fig:vo} visualizes this ablation on three scenes (\emph{Siren}, \emph{Pool}, \emph{Train}).

The \emph{All-Merge} variant merges all proposals belonging to the same category into a single instance, which oversmooths extended structures and loses important spatial variation (e.g., the long hedge in \emph{Siren} and the water surface in \emph{Pool}). The \emph{No-Merge} variant treats each proposal as a separate instance, often leading to excessive fragmentation (dozens of instances for a single physical object), which complicates downstream spatial audio allocation. Our voting-based strategy (\emph{Vote (Ours)}) aggregates proposals using semantic agreement while retaining a small number of coherent instances that better match human perception of sound-emitting regions. 

Moreover, \emph{Vote (Ours)} is more robust to failures introduced by splitting views. In \emph{Siren} and \emph{Pool}, certain regions are poorly captured in individual perspective tiles, \eg the tops of the bushes or the middle of the pool. These missing areas create artifacts for both \emph{All-Merge} and \emph{No-Merge}, while our voting scheme recovers them by relying on SAM2’s global panoramic proposals, which better preserve the overall scene structure. Finally, our merging strategy is agnostic to the particular 360$^\circ$ panorama grounding model used, and can readily benefit from future improvements in panoramic segmentation and grounding.

\subsection{Camera and Microphone Calibration}

Finally, we provide additional visualizations of the camera–microphone calibration process for \ourdataset. As discussed in the main paper, accurate alignment between the FOA microphone and the 360$^\circ$ camera is critical for learning reliable audio-visual correspondences.

\begin{figure*}
    \centering
    \includegraphics[width=\linewidth]{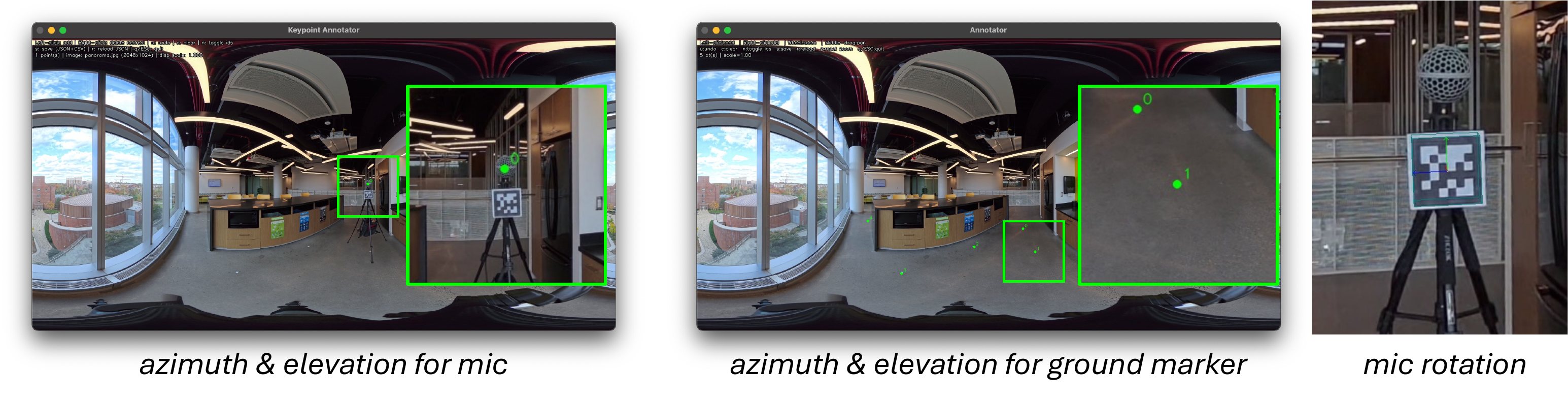}
    \caption{\textbf{Calibration annotation interface.}
    Our tool for camera–microphone calibration. Left: annotating the azimuth and elevation of the FOA microphone in the panorama. Middle: annotating azimuth and elevation for ground markers (\ie, the marker right below the microphone position) that define a world reference frame. Right: visualizing the relative rotation between AprilTag marker and camera, which is used to estimate the microphone orientation in the world coordinate system.}
    \label{fig:cali_kit}
\end{figure*}

Figure~\ref{fig:cali_kit} shows our annotation interface. In the left panel, annotators click on the microphone in the panorama to specify its azimuth and elevation relative to the camera. In the middle panel, they annotate the marker on the ground below the microphone. In the right panel, we visualize the estimated microphone rotation relative to the marker board. These annotations are combined with the AprilTag detections to obtain an initial estimate of the microphone pose.

\begin{figure*}
    \centering
    \includegraphics[width=\linewidth]{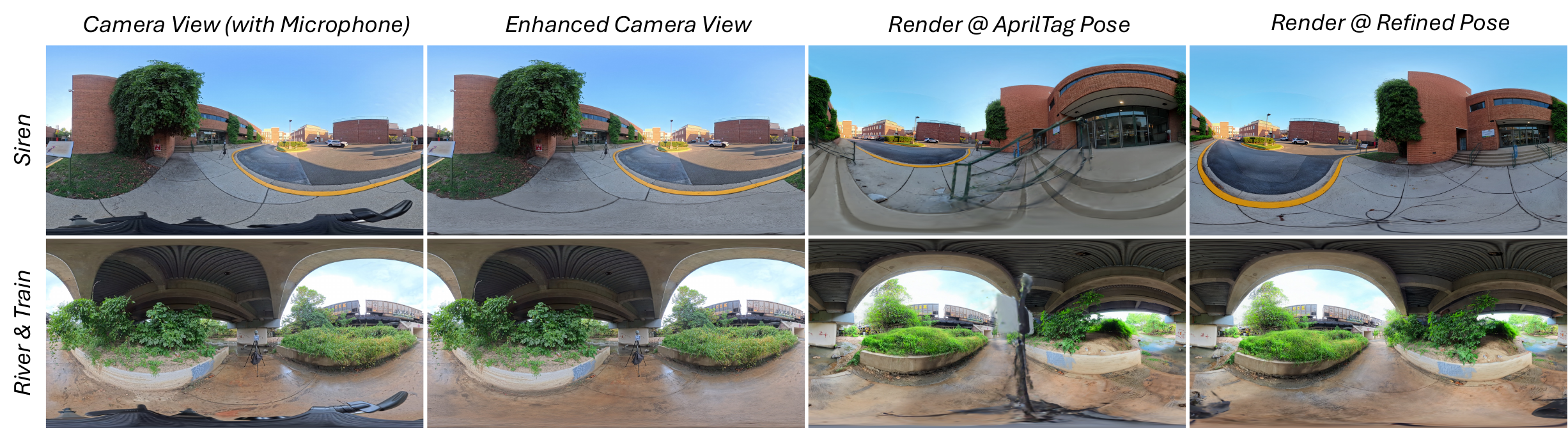}
    \caption{\textbf{Effect of calibration on rendered scenes.}
    For two real scenes (\emph{Siren} and \emph{River \& Train}), we show the original camera view with the microphone visible, the inpainted camera view, and renderings of the reconstructed 3D scene when aligned using the AprilTag-based pose estimate vs.\ our refined pose. The refined calibration produces renderings whose layout and perspective are more consistent with the original capture, leading to more accurate alignment between the FOA reference frame and the visual scene.}
    \label{fig:cali_refine}
\end{figure*}

Figure~\ref{fig:cali_refine} illustrates how the calibration affects the rendered views. For each scene, we show (i) the raw camera view with the microphone visible, (ii) an ``enhanced'' camera view where the microphone is removed via inpainting, (iii) the rendering of the reconstructed 3D scene when the FOA reference frame is aligned using only the AprilTag pose, and (iv) the rendering after our refinement. The refined poses yield better alignment between the camera and the inferred world frame, which provides reliable foundation for evaluation on \ourdataset.

\section{More Details For \ourdataset Dataset}

Remind that the input is one RGB image, and the outputs are:
\begin{itemize}
    \item 3D Visual Scene Representation (3DGS, for ours).
    \item Spatial Audio Field ${\bf A}$ (Ambisonics Field in our case)
\end{itemize}

\subsection{Hardware Setup and Microphone Calibration}

We briefly recall the \ourdataset setup and provide details about the calibration procedure (see Fig.~2 in the main paper). An Insta360 X5 camera captures 6K 360$^\circ$ video, and 12K 360$^\circ$ image for microphone calibration, and a R\O DE NT-SF1 FOA microphone records ambisonics at 48~kHz. The microphone is mounted with an AprilTag~\cite{apriltag} and ensured to be visible by the camera.

To use the FOA channels as a reference sound field aligned with the panorama, we must estimate:
\begin{itemize}
    \item the rotation between the microphone and the camera coordinate system, and
    \item the absolute azimuth and elevation of the microphone axes in the world frame.
\end{itemize}

\paragraph{Initial extrinsic calibration with AprilTag.}
We attach an AprilTag board~\cite{apriltag} to the microphone and capture a 12K image for calibration after the setup in a scene, and split the 360$^\circ$ image into 12 field-of-view (FOV) views. For each view, we:
\begin{enumerate}
    \item detect the AprilTag pose in the FOV image,
    \item reconstruct the board pose in the camera coordinate system, and
    \item choose the one with the smallest reconstruction error as final prediction.
\end{enumerate}
This yields an initial rotation quaternion $q_{\text{cam}\rightarrow\text{mic}}$ and translation offset.

\paragraph{Elevation and azimuth refinement.}
Due to inconsistent scale between reconstructed scale and real pose, we further refine the orientation using human annotation. For each scene, the annotator:
\begin{enumerate}
    \item selects a the middle of the microphone to annotate the elevation ($\theta_{\mathrm{mic}}$,$\varphi_{\mathrm{mic}}$)
    \item we stick some small markers on the ground to allow us annotate the point that is under the microphone, this will be use to query the depth and use to calculate the microphone in the generated scene, the reason we need another marker instead of using the microphone depth is that we found that the ground depth is usually more accurate/consistent than the microphone middle point.
\end{enumerate}
Combine with the depth estimated at the ground, we can 

\subsection{Dataset Statistics}

Table~\ref{tab:dataset_stats} reports scene-wise statistics for \ourdataset. For each scene, there might be more than one setup (e.g., \emph{Pool-left}/\emph{Pool-right}) under the scene categories described in the main paper (Fountain, Kitchen, Pool, Bridge, Stream, Siren), but we keep all subsets for evaluation.

\begin{table}[t]
    \centering
    \small
    \begin{tabular}{lccc}
        \toprule
        Scene (setup) & \# Mics & Clips / Mic & Total Clips \\
        \midrule
        Fountain              & 10 & 2 & 20 \\
        Kitchen               & 5  & 2 & 10 \\
        Pool-left             & 3  & 2 & 6  \\
        Pool-right            & 2  & 2 & 4  \\
        Bridge-river          & 1  & 2 & 2  \\
        Bridge-train          & 1  & 2 & 2  \\
        Stream-original       & 5  & 2 & 10 \\
        Stream-human     & 5  & 2 & 10 \\
        Building-siren                 & 1  & 1 & 1  \\
        Building-birds                 & 1  & 3 & 3  \\
        \midrule
        Total                 & 34 & -- & 68 \\
        \bottomrule
    \end{tabular}
    \caption{\textbf{Statistics of the \ourdataset dataset.} Each scene subset corresponds to one environment; we vary microphone layouts within a scene to capture diverse listening positions.}
    \label{tab:dataset_stats}
\end{table}

\section{Metric Definitions}

\subsection{Spatial Metrics}

\paragraph{DoA from FOA intensity.} Following~\cite{liu2025omniaudiogeneratingspatialaudio,Heydari2024ImmerseDiffusionAG}
for first-order ambisonics (FOA, $L=1$), we write the four channels as
\begin{equation}
    \mathbf{a}_1(t) = [W(t), Y(t), Z(t), X(t)]^\top,
\end{equation}
where $W$ is the omnidirectional channel and $(X,Y,Z)$ encode directional components in a right-handed coordinate system. We approximate the intensity vector by correlating $W$ with the directional channels over the clip:
\begin{align}
    I_X &= \sum_t W(t)\,X(t),\\
    I_Y &= \sum_t W(t)\,Y(t),\\
    I_Z &= \sum_t W(t)\,Z(t),
\end{align}
where $t$ indexes audio samples or short-time frames (we use STFT frames in practice). The intensity vector $\mathbf{I} = [I_X, I_Y, I_Z]^\top$ defines a dominant direction of arrival (DoA) on the unit sphere.

We convert $\mathbf{I}$ to azimuth $\theta$ and elevation $\varphi$:
\begin{align}
    \theta &= \operatorname{atan2}(I_Y, I_X),\\
    \varphi &= \operatorname{atan2}\big(I_Z,\ \sqrt{I_X^2 + I_Y^2}\big),
\end{align}
with $\theta\in[-\pi,\pi]$, $\varphi\in[-\frac{\pi}{2}, \frac{\pi}{2}]$.

Given ground-truth and predicted DoAs, $(\theta_{\mathrm{gt}}, \varphi_{\mathrm{gt}})$ and $(\theta_{\mathrm{pred}}, \varphi_{\mathrm{pred}})$, we report:
\begin{enumerate}
    \item 
\noindent\textbf{Azimuth error}
\begin{equation}
    \Delta_{\text{abs}}\theta
    =
    \min\big(
        |\theta_{\mathrm{gt}}-\theta_{\mathrm{pred}}|,
        2\pi - |\theta_{\mathrm{gt}}-\theta_{\mathrm{pred}}|
    \big)\,.
\end{equation}

\item\textbf{Elevation error}
\begin{equation}
        \Delta_{\text{abs}}\varphi
        =
        \big|\varphi_{\mathrm{gt}}-\varphi_{\mathrm{pred}}\big|.
\end{equation}

\item\textbf{Geodesic angular error.}
We convert both DoAs to points on the unit sphere,
\begin{align}
    \mathbf{u}_{\mathrm{gt}} &=
    \begin{bmatrix}
        \cos\varphi_{\mathrm{gt}}\cos\theta_{\mathrm{gt}}\\
        \cos\varphi_{\mathrm{gt}}\sin\theta_{\mathrm{gt}}\\
        \sin\varphi_{\mathrm{gt}}
    \end{bmatrix},\quad
    \mathbf{u}_{\mathrm{pred}} =
    \begin{bmatrix}
        \cos\varphi_{\mathrm{pred}}\cos\theta_{\mathrm{pred}}\\
        \cos\varphi_{\mathrm{pred}}\sin\theta_{\mathrm{pred}}\\
        \sin\varphi_{\mathrm{pred}}
    \end{bmatrix},
\end{align}
and compute the geodesic distance
\begin{equation}
    \Delta_{\text{Angular}}
    =
    \arccos\big(
        \operatorname{clip}(
            \mathbf{u}_{\mathrm{gt}}^\top\mathbf{u}_{\mathrm{pred}}, -1, 1
        )
    \big).
\end{equation}
For completeness, we also report the haversine-style formulation used in the main text:
\begin{gather}
    a = \sin^2\left(\frac{\Delta_{\mathrm{abs}}\theta}{2} \right)
    + \cos\varphi_{\mathrm{pred}}\cos\varphi_{\mathrm{gt}}
    \sin^2\left(\frac{\Delta_{\mathrm{abs}}\varphi}{2}\right),\\[0.25em]
    \Delta_{\text{Angular}} = 2\arctan\left(\sqrt{\frac{a}{1-a}}\right).
\end{gather}

\end{enumerate}

\paragraph{Spherical energy maps, CC and AUC.}

Following~\cite{visage}, given ambisonics coefficients $a_L(t)$, we render the scalar pressure at a direction $(\theta,\varphi)$ as
\begin{equation}
    a(\theta,\varphi,t) = y_L(\theta,\varphi)^\top a_L(t).
\end{equation}
We then compute a time-aggregated energy map
\begin{equation}
    E(\theta,\varphi) = \sum_t |a(\theta,\varphi,t)|^2,
\end{equation}
discretized on a balanced spherical grid
$\Omega = \{(\theta_i,\varphi_i)\}_{i=1}^N$ (we use an equiangular grid with uniform weights).

Let $\mathbf{e}_{\mathrm{pred}}, \mathbf{e}_{\mathrm{gt}}\in\mathbb{R}^N$ be the flattened predicted and ground-truth energy maps after min--max normalization to $[0,1]$. The \emph{correlation coefficient} (CC) is
\begin{equation}
    \mathrm{CC} =
    \frac{
        \sum_i (\mathbf{e}_{\mathrm{pred},i} - \bar{e}_{\mathrm{pred}})
               (\mathbf{e}_{\mathrm{gt},i} - \bar{e}_{\mathrm{gt}})
    }{
        \sqrt{\sum_i (\mathbf{e}_{\mathrm{pred},i} - \bar{e}_{\mathrm{pred}})^2}
        \sqrt{\sum_i (\mathbf{e}_{\mathrm{gt},i} - \bar{e}_{\mathrm{gt}})^2}
    }.
\end{equation}

To compute AUC, we treat $\mathbf{e}_{\mathrm{gt}}$ as a soft foreground mask by binarizing it at its median value. Using $\mathbf{e}_{\mathrm{pred}}$ as scores, we form the ROC curve over all thresholds and report the area under the curve (AUC), following ViSAGe~\cite{visage}. Higher CC and AUC indicate closer spatial energy patterns to the reference.

\subsection{Semantic Metrics}

\paragraph{Directional CLAP.}
We evaluate semantic consistency by probing the ambisonics field along four canonical FOA-aligned directions:
\begin{align}
    \text{left:}\quad &\mathbf{u}_\text{L} = (\theta=\tfrac{\pi}{2}, \varphi=0),\\
    \text{right:}\quad &\mathbf{u}_\text{R} = (\theta=-\tfrac{\pi}{2}, \varphi=0),\\
    \text{front:}\quad &\mathbf{u}_\text{F} = (\theta=0, \varphi=0),\\
    \text{back:}\quad &\mathbf{u}_\text{B} = (\theta=\pi, \varphi=0).
\end{align}
For each direction $\mathbf{u}_d$, we render a monaural waveform
\begin{equation}
    a_d(t) = y_L(\mathbf{u}_d)^\top a_L(t).
\end{equation}

Let $f_a(\cdot)$ and $f_t(\cdot)$ be the audio and text encoders of CLAP~\cite{laionclap2023}. For a caption $c$ describing a sounding source and directional audio $a_d$, we define
\begin{align}
    s_{\mathrm{CLAP-T}}(d, c) &= \cos\big(f_a(a^{\mathrm{pred}}_d), f_t(c)\big),\\
    s_{\mathrm{CLAP-A}}(d)    &= \cos\big(f_a(a^{\mathrm{pred}}_d),
                                         f_a(a^{\mathrm{gt}}_d)\big),
\end{align}
where $\cos(\cdot,\cdot)$ denotes cosine similarity. We report:
\begin{itemize}
    \item \textbf{D-CLAP$_\text{T}$}: averaged $s_{\mathrm{CLAP-T}}$ between directional audio and its text caption;
    \item \textbf{D-CLAP$_\text{A}$}: averaged $s_{\mathrm{CLAP-A}}$ between predicted and ground-truth directional audio;
    \item \textbf{D-CLAP$_\text{R}$}: for each annotation $(d_{\mathrm{gt}}, c)$, we rank all four directions by $s_{\mathrm{CLAP-T}}(d, c)$ and compute top-1 accuracy that $d_{\mathrm{gt}}$ is ranked highest.
\end{itemize}
All metrics are averaged over clips in \ourdataset.

\section{More Details For \ourframework Framework}

\subsection{VLM Prompt}
We query a vision-language model (VLM) with the input image $I$ to obtain a list of sounding categories, their types (point/cluster/global), and audio prompts for text-to-audio generation. In Figure~\ref{fig:vlm-prompt-full} we provide an example prompt used in practice.

\begin{figure*}[t]
  \centering
  \begin{scriptsize}
  \begin{verbatim}
TASK
You are given ONE image. Infer plausible sounds from VISIBLE things in the image.

OUTPUT
Return a JSON array with 2–8 items, ordered from loudest (peak_db closest to 0) to softest (most negative). 
Output JSON ONLY—no extra text.

FOR EACH ITEM (object)
- "diffusion_prompt": <=10 simple words describing the sound (common words only).
- "grounding_label": 1–2 word VISIBLE object to ground on (e.g., river, tree, door).
  - If the sound comes from a hidden agent (bird, insect, person off-frame), map it to a VISIBLE HOST object 
    (e.g., tree, bush, window, street).
- "peak_db": integer NEGATIVE dB for target PEAK (0 dB = full-scale, where [-1, 1] is 0 dB). Never use values > -6 dB.
- "source_type": Choose from "area", "point", and "background", area means the sound comes from an area, 
    e.g., river, leaves, "point" means the sound comes from a point source, 
    "background" is reserved for global background bed.

GLOBAL BACKGROUND BED (required as LAST item)
- Add ONE final item that captures the scene’s background as a reusable, loopable bed: 
    e.g., "open room soft air hum", "damp cave low drip hush", "quiet library room tone", 
    or "silence" if none is implied.
- Use "grounding_label": "global".
- Set "peak_db" soft (\approx -26...-32). If truly silent, use -120.
- Set "source_type" as "background"

SELECTION RULES
- Include only sources that are visible or strongly implied by what is visible 
    (moving water -> water sound; swaying trees -> wind; visible vents -> HVAC).
- No voices/music unless people/speakers are visible.
- Keep words simple; avoid jargon, metaphors, and long hyphen chains.
- Use Ascii.

LEVEL GUIDE (choose by strength)
- Foreground/strong: -8...-14 dB
- Mid/medium: -14...-20 dB
- Far/quiet: -20...-26 dB
- Background bed: -26...-32 dB (or -120 for silence)

CONSTRAINTS
- Keys must be exactly: "diffusion_prompt", "grounding_label", "peak_db", "source_type".
- Use integers for "peak_db".
- "source_type" options: area, point, background
- Do not add other fields.
OUTPUT-TEMPLATE
- Output in JSON format:
```
 [
  {"diffusion_prompt": "<prompt>", "grounding_label": "<object>", "peak_db": <int>, "source_type": <area/point>},
  {"diffusion_prompt": "<prompt>", "grounding_label": "<host object>", "peak_db": <int>", source_type": <area/point>},
  {"diffusion_prompt": "<generative background of the scene>", "grounding_label": "global", "peak_db": <int>}
]
```
  \end{verbatim}
  \end{scriptsize}
  \caption{VLM prompt used to query the vision-language model.}
  \label{fig:vlm-prompt-full}
\end{figure*}
\begin{verbatim}

\end{verbatim}

The JSON output is parsed and converted into the category set $\mathcal{C}$ and equalization parameters used by our spatial audio encoder.

The warping operator $W_G$ reprojects a calibrated perspective image into an equirectangular panorama while avoiding aliasing near the poles and at large viewing angles.

\paragraph{Mapping from panorama to camera.}
For each output panorama pixel $(u,v)$ with resolution $W_\text{pano}\times H_\text{pano}$, we convert to spherical angles
\begin{align}
    \theta &= 2\pi\left(\frac{u+0.5}{W_\text{pano}} - \tfrac{1}{2}\right),\\
    \varphi &= \pi\left(\frac{v+0.5}{H_\text{pano}} - \tfrac{1}{2}\right),
\end{align}
and obtain a direction vector
\begin{equation}
    \mathbf{d}(\theta,\varphi) =
    \begin{bmatrix}
        \cos\varphi\cos\theta\\
        \cos\varphi\sin\theta\\
        \sin\varphi
    \end{bmatrix}.
\end{equation}

\subsection{Gaussian-Pyramid-Based Warping}

Given the camera extrinsics from GeoCalib~\cite{veicht2024geocalib}, we transform $\mathbf{d}$ into the camera frame and project to normalized image coordinates $(x,y)$ using the calibrated focal length $f$ from Eq. (4) in the main text. These are mapped to pixel coordinates $(u',v')$ in the input image.

\paragraph{Gaussian pyramid and anti-aliased sampling.}
Foreshortening near the panorama poles and along the vertical direction leads to non-uniform sampling: some panorama pixels correspond to large footprints in the input image. To mitigate aliasing, we construct a Gaussian pyramid $\{I^{(s)}\}_{s=0}^{S-1}$ from the input image $I$, where $I^{(0)}=I$ and
\begin{equation}
    I^{(s+1)} = \mathrm{downsample}_2(\mathrm{GaussianBlur}(I^{(s)})).
\end{equation}
For each panorama pixel, we approximate a local magnification factor $\rho$ from the Jacobian of the equirectangular-to-camera mapping and choose a pyramid level
\begin{equation}
    s^\star = \mathrm{clip}\!\left(\left\lfloor \log_2 \rho \right\rceil, 0, S-1\right),
\end{equation}
where $\lfloor\cdot\rceil$ denotes rounding to the nearest integer. We then sample $I^{(s^\star)}$ at $(u',v')$ using bilinear interpolation to obtain the warped color. This yields
\begin{equation}
    I_{\text{warp}} = \mathcal{W}_G(I,\varphi,f),
\end{equation}
which is used as input to the panorama outpainting model $g_{\text{outpaint}}$.

\paragraph{Voting}
We address the mismatch between FoV-tile-wise open-vocabulary segmentation (OVS) masks and globally consistent panoramic masks by letting SAM2 \cite{ravi2024sam2} proposals and X-Decoder \cite{xdecoder} instance masks vote via Alg.~\ref{alg:mask_voting_algo}, preserving SAM2’s global geometry while inheriting X-Decoder’s category-wise semantics.

For each SAM2 proposal $\mathcal{M}^{\text{pano}}_i$ and candidate sounding category $c$ proposed by GPT-5~\cite{gpt5}, Alg.~\ref{alg:mask_voting_algo} computes a vote score $s_i$ by aggregating per-pixel confidences from overlapping open-vocabulary masks $\mathbf{M}_{\text{OVS},c}$. 
\textsc{PixelScore}$(p)$ assigns to each pixel $p$ in $\mathcal{M}^{\text{pano}}_i$ the maximum confidence over all instance masks in $\mathbf{M}_{\text{OVS},c}$ whose IoU with $\mathcal{M}^{\text{pano}}_i$ exceeds $\tau_{\text{IOU}}$. 
\textsc{Visibility}$\bigl(\mathcal{M}^{\text{pano}}_i,\mathbf{M}_{\text{OVS},c}\bigr)$ counts the pixels in $\mathcal{M}^{\text{pano}}_i$ that receive at least one such vote and normalizes the sum of \textsc{PixelScore}$(p)$ to obtain $s_i$. 
If $s_i \ge \tau_{\text{vote}}$, \textsc{Combine}$\bigl(\mathcal{M}^{\text{pano}}_i,\mathbf{M}_{\text{OVS},c}\bigr)$ takes the union of all contributing instance masks (those with $\text{IoU} > \tau_{\text{IOU}}$) with $\mathcal{M}^{\text{pano}}_i$ to produce the refined mask $\widehat{\mathcal{M}}_i$, which is then added to $\mathbf{M}_c$, the final set of retained and refined panoramic instance masks for category $c$.

\begin{algorithm}[t]
\caption{Mask Voting for Category $c$}
\label{alg:mask_voting_algo}
\begin{algorithmic}[1]
\State \textbf{Input:} $\mathbf{M}_{\text{pano}},\,\mathbf{M}_{\text{OVS},c},\,\tau_{\text{vote}},\,\tau_{\text{\text{IOU}}}$
\State \textbf{Output:} $\mathcal{M}_c$
\State $\mathbf{M}_c \gets \emptyset$
\For{each $\mathcal{M}^{\text{pano}}_i \in \mathbf{M}_{\text{pano}}$}
    
    \State $s_i \gets \frac{\sum_p \textsc{PixelScore}(p)}{\textsc{Visibility}(\mathcal{M}^{\text{pano}}_i, \mathbf{M}_{\text{OVS},c})}$
    \If{$s_i \ge \tau_{\text{vote}}$}
        \State $\widehat{\mathcal{M}}_i \gets \mathcal{M}^{\text{pano}}_i
               \lor \textsc{Combine}\bigl(\mathcal{M}^{\text{pano}}_i,\mathcal{M}_{\text{OVS},c}\bigr)$
        \State $\mathbf{M}_c \gets \mathbf{M}_c \cup \{\widehat{\mathcal{M}}_i\}$
    \EndIf
\EndFor
\State \textbf{Return} $\mathbf{M}_c$
\end{algorithmic}
\end{algorithm}

\paragraph{Point Downsampling}
Given a semantic mask $\mathcal{M}_i$ for object $i$ in the equirectangular panorama and its associated 3D point set $\mathcal{P}_{\text{raw,}i} = \{\mathbf{x}_j\}_{j=1}^{N_i}$, where each point has elevation $e_j$, depth $d_j$, surface normal $\mathbf{n}_j$, and view direction $\mathbf{v}_j$, we compute per-point importance weights to form a compact set of representatives while preserving extended geometry:
\begin{equation}
w_j = \frac{d_j^2\cos e_j}{ \max\left(|\mathbf{n}_j^\top \mathbf{v}_j|, \varepsilon\right)}\,,
\label{eq:weights}
\end{equation}
where the $\cos e_j$ term compensates for equirectangular sampling density, $d_j^{2}$ implements distance-based weigthing, and the normal term deprioritizes visible surfaces, and $\varepsilon$ is a small number for numerical stability. 
We normalize $\pi_j = w_j / \sum_{\ell=1}^{N_i} w_\ell$ and perform weighted down-sampling to at most $N_{\max}=1000$ representatives, denote $\mathcal{P}_i$ as the final down-sampled point cloud of object $i$ and $\mathbf{x}_{ik}$ as the $k^{\text{th}}$ point in ${\mathcal P }_i$:
\begin{equation}
\mathcal{P}_i = \{\mathbf{x}_{ik}\}_{k=1}^{K_i}, \qquad 
\mathbf{x}_{ik} \sim \mathcal{P}_{\text{raw,}i}
\label{eq:sampling}
\end{equation}
where $\Pr(\mathbf{x}_j)=\pi_j$ and $K_i =\min(N_{\max},N_i)$.

\subsection{From Directional Field to Binaural Audio}

We elaborate on the HRTF-based binaural decoding used in Sec.~4.4 of the main paper. Given the directional sound field $a(\theta,\varphi,t)$ and left/right head-related impulse responses (HRIRs) $h_{\text{left}}(\theta,\varphi,\tau)$ and $h_{\text{right}}(\theta,\varphi,\tau)$, the binaural signals can be written as
\begin{align}
    b_{\text{left}}(t) &= \sum_{\theta,\varphi}\sum_{\tau}  \big(h^{\text{left}}(\theta,\varphi,\tau) \cdot a(\theta,\varphi,t-\tau)\big) ,\\
    b_{\text{right}}(t) &= \sum_{\theta,\varphi}\sum_{\tau} \big(h^{\text{right}}(\theta,\varphi,\tau) \cdot a(\theta,\varphi,t-\tau)\big),
\end{align}
where the sum is over a discrete sampling of the sphere.

Using the ambisonics expansion (Eq. (1) in main)
\begin{equation}
    a(\theta,\varphi,t) = \sum_{\ell,m} Y_\ell^m(\theta,\varphi)\,a_{\ell,m}(t),
\end{equation}
we can precompute ambisonics-domain HRIRs:
\begin{equation}
    h^{\text{left/right}}_{\ell,m}(\tau)
    = \sum_{\theta,\varphi} w(\theta,\varphi)\,
      h^{\text{left/right}}(\theta,\varphi,\tau)\,Y_\ell^m(\theta,\varphi),
\end{equation}
where $w(\theta,\varphi)$ are weights for spherical integration. Substituting into the binaural equations yields
\begin{equation}
    \begin{bmatrix}
        b_{\text{left}}(t)\\
        b_{\text{right}}(t)
    \end{bmatrix}
    =
    \sum_{\ell=0}^L\sum_{m=-\ell}^{\ell}
    \begin{bmatrix}
        h^{\text{left}}_{\ell,m} * a_{\ell,m}\\[0.25em]
        h^{\text{right}}_{\ell,m} * a_{\ell,m}
    \end{bmatrix}(t),
\end{equation}
which matches Eq.~(11) in the main paper after stacking channels into vectors. In practice, we precompute $h^{\text{left}}_{\ell,m}$ and $h^{\text{right}}_{\ell,m}$ by a regular HRTF set (\eg, SADIE-II~\cite{sadie2} dataset).

\section{Setup for One-Shot Room Acoustic Learning}

We expand on Sec.~5.4 of the main paper. Here, the goal is to fit the acoustic parameters of our differentiable renderer so that the predicted ambisonics match a \emph{single} first-order ambisonics (FOA) recording at one microphone pose.

\paragraph{Task formulation.}
Let $\tilde{a}_L(t)$ denote the ground-truth FOA signal at a fixed pose $\tilde{\mathbf{p}}$, and $a_{\mathrm{src}}(t)$ be the monaural dry source audio. Let ${\bf A}(\mathbf{p},t;\theta)$ be our renderer with learnable parameters $\theta$, including:
\begin{enumerate}[label=(\roman*)]
    \item the attenuation constant $\alpha$ controlling geometric decay,
    \item the average frequency-dependent reflection response $R[f]$ of the room surfaces,
    \item per-source equalization coefficients $s$ (gain and tilt),
    \item and the predicted RT60 $\hat{T}_{60}$ (in seconds), parameterized as a base RT60, by $\rho$ and a frequency slope, by $\gamma$.
\end{enumerate}
The rendered FOA sound field at listener pose $\mathbf{p}$ is
\begin{align}
    {\bf A}(\mathbf{p}, t; \theta)
    &= \big[\mathrm{RIR}_L(\mathbf{p}, \cdot;\theta) * a_{\text{src}}(\cdot)\big](t),
\end{align}
where $\mathrm{RIR}_L(\mathbf{p}, t;\theta)\in\mathbb{R}^{(L+1)^2}$ is the ambisonic room impulse response that models the transfer function between the source and $\mathbf{p}$. Following~\cite{Jin_2025_ICCV,hearinganythinganywhere2024}, we decompose it into early reflections and late diffuse reverberation:
\begin{align}
    \mathrm{RIR}_L(\mathbf{p}, t;\theta)
    &= \mathrm{Blend}[\mathrm{RIR}^{\text{early}}_L(\mathbf{p}, t;\theta),
      \mathrm{RIR}^{\text{late}}_L(\mathbf{p}, t;\theta)].
\end{align}

The early part is modeled as a sum over geometric paths $p\in\mathbb{P}$ (direct path and a sparse set of early reflections from beam tracing~\cite{Jin_2025_ICCV,beam1,beam2,beam3,beam4}):
\begin{align}
    &\mathrm{RIR}^{\text{early}}_L(\mathbf{p}, t;\theta)\\
    =& \frac{se^{-\alpha t}}{c_{\text{sound}}\tau_p} \sum_{p\in\mathbb{P}}
        \mathbf{y}_L(\mathbf{d}_p)\;
        \mathcal{F}_{\min}^{-1}\!\left\{
            R[f]^{|p|}
        \right\}(t - \tau_p),
\end{align}
where $\mathbf{y}_L(\mathbf{d}_p)$ is the ambisonic encoding of the path ending direction $\mathbf{d}_p$, $|p|$ is the number of reflections along path $p$, $\tau_p$ is the path delay, and $\mathcal{F}_{\min}^{-1}$ denotes min-phase transform. The reflection response $R[f]$ and equalization $s$ control the spectral characteristics of early reflections, while $\alpha$ governs distance-dependent attenuation modeling air aborption.

\paragraph{Late reverberation parameterization from RT60.}
The late part $\mathrm{RIR}^{\text{late}}_L$ captures the dense, diffuse tail beyond the early reflection window. We approximate it using a stochastic, frequency-dependent exponential decay that is fully determined by a small number of RT60 parameters.

In our implementation, the late tail is first synthesized as a mono signal $r^{\text{late}}(t;\theta)$ and then mapped to ambisonics under a diffuse-field assumption (i.e., equal energy in all directions). For each band $b$ we construct an exponentially decaying envelope
\begin{equation}
    e_b(t) = \exp\!\left(
        -\frac{\ln(1000)}{\hat{T}_{60}(f_b)}\, t
    \right),
\end{equation}
where $\ln(1000)$ corresponds to a 60\,dB decay (i.e., the definition of RT60). The band-wise late reverberation signals are then
\begin{equation}
    r_b(t;\theta) = e_b(t)\,\tilde{n}_b(t),
\end{equation}
and we sum across bands to obtain a full-band late tail
\begin{equation}
    r^{\text{late}}(t;\theta)
    = \sigma(g)\sum_{b=1}^B r_b(t;\theta),
\end{equation}
where $g$ is a learnable gain parameter and $\sigma(\cdot)$ is a sigmoid to keep the overall late tail level bounded and stable. Finally, we normalize $r^{\text{late}}$ to have unit peak magnitude during training.

\paragraph{From mono tail to ambisonics.}
Under a diffuse-field assumption, late reverberation is approximately isotropic. We therefore lift the mono late tail $r^{\text{late}}(t)$ to ambisonics by:
\begin{equation}
    \mathrm{RIR}^{\text{late}}_L(\mathbf{p}, t;\theta)
    = r^{\text{late}}(t;\theta)\,\mathbf{1},
\end{equation}
where $\mathbf{1}\in\mathbb{R}^{(L+1)^2}$ broadcase $r^{\text{late}}$ to all channels. This gives a spatially smooth, low-variance late tail that complements the directional early reflections.

\paragraph{Early/late blending.}
We model the full room impulse response as a smooth combination of a deterministic early part and a stochastic late tail. Intuitively, the early reflections (direct path and a few specular bounces) encode precise geometric information, while the late reverberation behaves more like a diffuse sound field. To avoid audible discontinuities between these two regimes, we introduce a time-dependent blend:
\begin{align}
    &\mathrm{RIR}_L(\mathbf{p}, t;\theta)\\
    =& w_{\text{early}}(t)\,
      \mathrm{RIR}^{\text{early}}_L(\mathbf{p}, t;\theta)
    + w_{\text{late}}(t)\,
      \mathrm{RIR}^{\text{late}}_L(\mathbf{p}, t;\theta),
\end{align}
where $w_{\text{early}}(t)$ and $w_{\text{late}}(t)$ are scalar envelopes that satisfy
\[
    w_{\text{early}}(t) \approx 1,\ w_{\text{late}}(t) \approx 0
    \quad\text{at very early times,}
\]
and
\[
    w_{\text{early}}(t) \approx 0,\ w_{\text{late}}(t) \approx 1
    \quad\text{well into the late tail.}
\]

We choose a physically motivated early/late cutoff time $T_{\mathrm{e}}$ (proportional to the window used to define early reflections) and construct a short cross-fade region around $T_{\mathrm{e}}$. Before this region, $w_{\text{early}}(t)$ stays close to one and $w_{\text{late}}(t)$ stays close to zero; after the cutoff, the roles are reversed. In the transition interval, we use a smooth cosine-shaped cross-fade so that both envelopes vary continuously and the total energy does not exhibit sharp jumps.

Finally, the overall level of the late tail is normalized relative to the early part: we set the initial amplitude of $\mathrm{RIR}^{\text{late}}_L$ such that its peak is comparable to the peak energy of $\mathrm{RIR}^{\text{early}}_L$ per ambisonics channel. This ensures a perceptually continuous decay from the last prominent early reflection into the diffuse reverberant tail, while still allowing the late component to adapt its decay rate and spectral color through the RT60-based parameterization described above.

\paragraph{One-shot fitting objective.}
Given the dry source $a_{\mathrm{src}}(t)$ and measured FOA $\tilde{a}_L(t)$ at pose $\tilde{\mathbf{p}}$, we optimize:
\begin{equation}
    \mathcal{L}(\theta)
    = \mathcal{L}_{\text{MAG}}
\end{equation}
And evaluated on $\mathcal{L}_{\text{MAG}}$, $\mathcal{L}_{\text{ENV}}$ and $\Delta_{\text{Angular}}$
where:
\begin{align}
    \mathcal{L}_{\text{MAG}} &=
    \big\|\log |S(A(\tilde{p},t;\theta))|
          - \log |S(\tilde{a}_L(t))|\big\|_1,\\
    \mathcal{L}_{\text{ENV}} &=
    \big\|\mathrm{Env}(A(\tilde{p},t;\theta))
          - \mathrm{Env}(\tilde{a}_L(t))\big\|_2^2,
\end{align}
$S(\cdot)$ is the STFT, $\mathrm{Env}(\cdot)$ computes the Hilbert-envelope per channel, and $\Delta_{\text{Angular}}$ is the geodesic angular error defined earlier. We optimize $\theta$ with Adam for a small number of iterations (one-shot setting), while keeping the 3D geometry and semantic anchors fixed.

\section{Setup for Audio-Visual Spatial Source Separation}

We also treat our renderer as a differentiable spatialization module for audio-visual source separation.

\paragraph{Mixture model.}
Given:
\begin{itemize}
    \item visually localized sources with 3D anchors, and
    \item a mixture FOA recording $\tilde{a}_L(t)$ at pose $\tilde{p}$,
\end{itemize}
we seek per-source monaural signals $\{s_i(t)\}_{i\in\mathcal{O}}$ such that
\begin{equation}
    \sum_{i\in\mathcal{O}} {\bf A}_i(\tilde{p},t;s_i) \approx \tilde{a}_L(t),
\end{equation}
where ${\bf A}_i$ denotes the contribution of source $i$ under our encoder in main Sec.~4.3.

\paragraph{Optimization objective.}
We parameterize $s_i$ either directly as learnable waveforms constrained by audio priors, or as latent codes for a text-to-audio prior that we decode at each iteration. The loss combines reconstruction and spatial regularization:
\begin{equation}
    \mathcal{L}_{\text{sep}} =
    \mathcal{L}_{\text{MAG}}(\textstyle\sum_i {\bf A}_i, \tilde{a}_L),
\end{equation}
where $\mathcal{L}_{\text{MAG}}$ is as above. In practice, we implement separation via diffusion posterior sampling~\cite{dps} over the per-source latents, guided by $\mathcal{L}_{\text{sep}}$.

\paragraph{Source Separation by Diffusion Posterior Sampling~\cite{dps}}

In our final model, we adopt a generative approach: each $s_i$ is sampled from a pretrained text-to-audio diffusion prior conditioned on the visual and textual description of source $i$, and the renderer acts as a differentiable observation model that ties all sources together through the FOA mixture.

Let $x^{(i)}_t$ denote the noisy latent of source $i$ at reverse-diffusion time step $t$, and let $p_{\text{prior}}(x^{(i)}_t)$ be the corresponding prior distribution given by the pretrained diffusion model. The posterior over latents given the observed mixture is
\begin{equation}
    p_{\text{post}}\big(\{x^{(i)}_t\}_{i\in\mathcal{O}} \,\big|\, \tilde{a}_L\big)
    \propto
    \Bigg[
        \prod_{i\in\mathcal{O}} p_{\text{prior}}(x^{(i)}_t)
    \Bigg]
    \exp\big(-\lambda\,\mathcal{L}_{\text{sep}}\big),
\end{equation}
where $\lambda$ controls the strength of the guidance.

During sampling, we approximate the posterior score for each source $j$ as
\begin{align}
    &\nabla \log p_{\text{post}}\big(x^{(j)}_t\big)\\
    \approx&
    \nabla \log p_{\text{prior}}\big(x^{(j)}_t\big)
    - \lambda\,\nabla_{x^{(j)}_t}
    \mathcal{L}_{\text{sep}}\!\Big(
        \{\mathbf{A}_i(\tilde{\mathbf{p}}, t; x^{(i)}_t)\},
        \tilde{a}_L(t)
    \Big),
\end{align}
where the second term backpropagates the separation loss through the renderer and the text-to-audio decoder into the latent of source $j$. This \emph{posterior-guided} score replaces the unconditional prior score in the reverse diffusion update, yielding a \emph{diffusion posterior sampler} that steers each source towards waveforms that (i) remain likely under the pretrained prior and (ii) jointly reconstruct the observed FOA mixture with spatial patterns consistent with the 3D audio-visual scene.

Intuitively, the pretrained diffusion model maintains the naturalness and diversity of individual source signals, while the renderer and $\mathcal{L}_{\text{sep}}$ enforce that their spatialized superposition matches the recorded sound field and respects the visual layout.

\paragraph{Initialization.}
We initialize all sources in a simple, physically motivated way. For each source, we first pan the mixture FOA into a directional mono signal according to Eq.~(1) in the main paper, using the angle predicted by the grounding model for that source. We then apply ZeroSep~\cite{huang2025zerosep} to enhance this directional signal and use the result as the initial waveform. Finally, starting from an intermediate diffusion time $t=0.5$, we run diffusion posterior sampling on Stable Audio Open~\cite{evans2024stableaudioopen} using the \texttt{DPM++(3M) SDE} solver.

\section{User Study Details}
\begin{figure}[t!]
    \centering
    \includegraphics[width=1\linewidth]{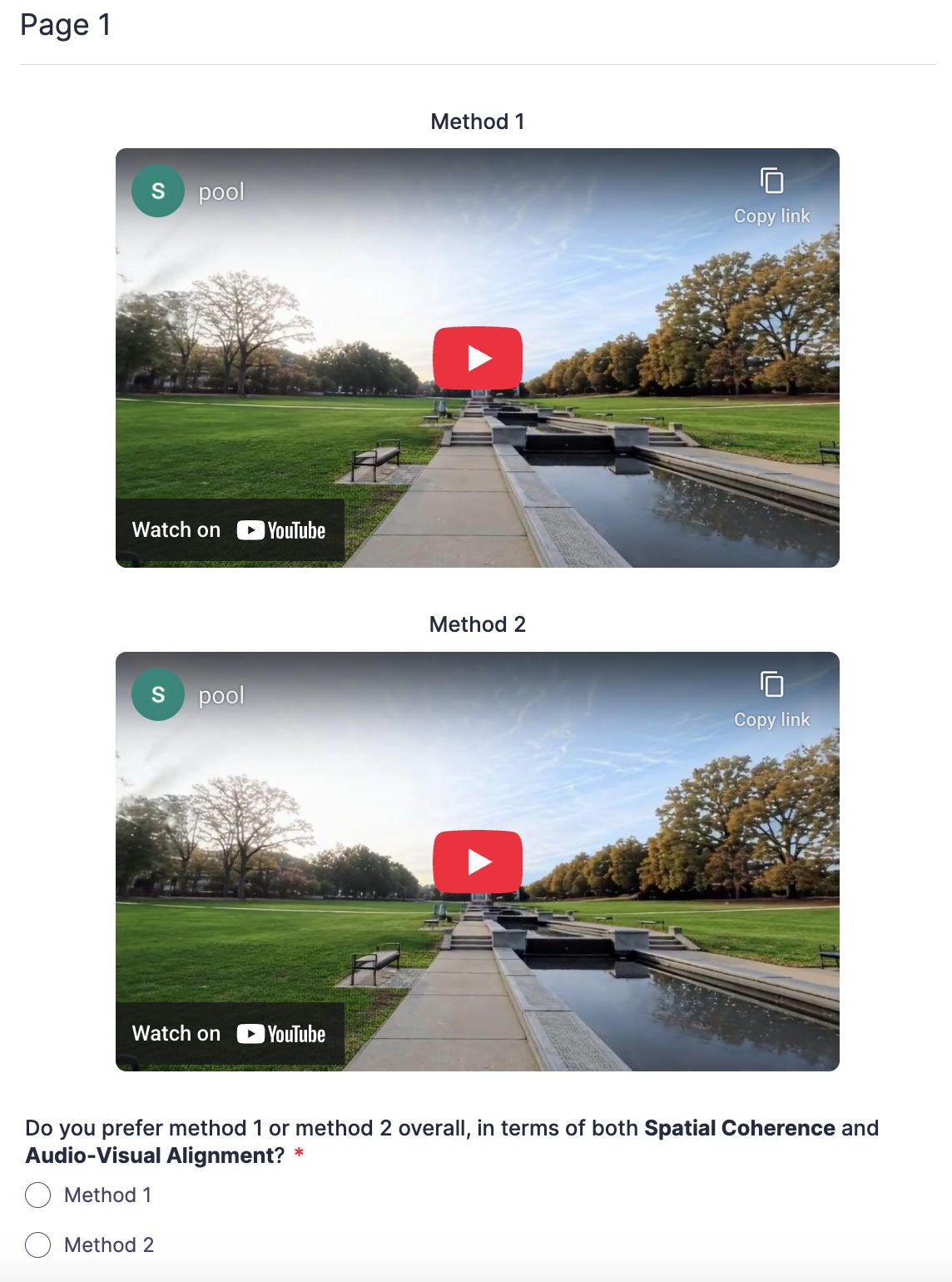}
    \caption{\textbf{User Study Interface.} An example pairwise comparison shown to participants. Each question presents two spatial-audio videos (“Method 1” and “Method 2”) for the same scene. Participants listen with headphones and select the method that provides better spatial coherence and audio–visual alignment.}
    \label{fig:user_study_interface}
\end{figure}

We conduct an anonymized user study to evaluate the spatial audio generation quality of our method in comparison to MMAudio~\cite{mmaudio} and OmniAudio~\cite{liu2025omniaudiogeneratingspatialaudio}. We recruit 50 participants via Prolific, selecting AI Taskers from diverse geographic regions. The study is administered through Zoho Forms. For each of the 12 test scenes, we form three pairwise comparisons: Ours vs. MMAudio, Ours vs. OmniAudio, and OmniAudio vs. MMAudio. This yields a total of 36 questions, which are presented in randomized order with the method names hidden to avoid biases. Before beginning, participants read a detailed explanation of the evaluation criteria and are instructed to judge each pair based on spatial coherence and audio–visual alignment. To ensure proper perception of binaural audio, participants must also confirm that they are wearing headphones prior to proceeding. Figure~\ref{fig:user_study_interface} shows an example question from our user study form.

\section{Discussions}

\noindent\textbf{Outdoor-focused propagation and reverberation.}
Our propagation model is designed primarily for outdoor scenes, where recordings are often close to dry~\cite{traer2016} and listeners generally have weaker expectations of strong reverberation~\cite{TRAER2021}. Since MMAudio is trained on web videos that typically exhibit little room-like reverberation, its outputs for common outdoor sources are also usually near-dry. As a result, double reverberation is unlikely in our setting. We therefore focus on the central challenges of semantic coherence, accurate DoA alignment, and heterogeneous source decomposition.

\begin{figure}[t!] 
    \centering
    \includegraphics[width=\linewidth]{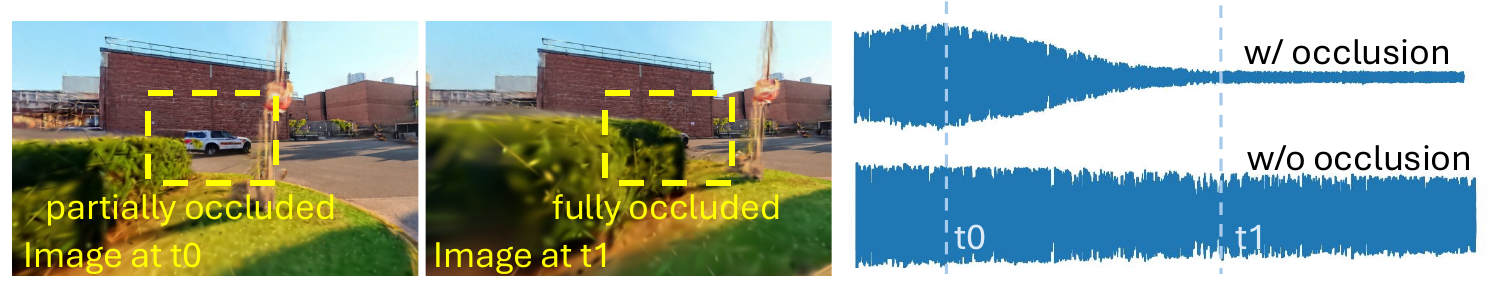}
    \caption{Physics-aware occlusion add-on: as a clustered source becomes partially and then fully blocked between $t_0$ and $t_1$, visibility-based point reweighting yields smooth attenuation of the rendered signal.}
    \label{fig:occlusion}
\end{figure}

\noindent\textbf{Physics-aware extensions.}
\ourframework naturally accommodates lightweight geometry-driven extensions. As one example, we incorporate a simple occlusion heuristic for clustered sources based on visibility-aware point reweighting, which yields smooth attenuation as a source transitions from partially visible to fully occluded (Fig.~\ref{fig:occlusion}).

\begin{figure}[t!]
    \centering
    \includegraphics[width=\linewidth]{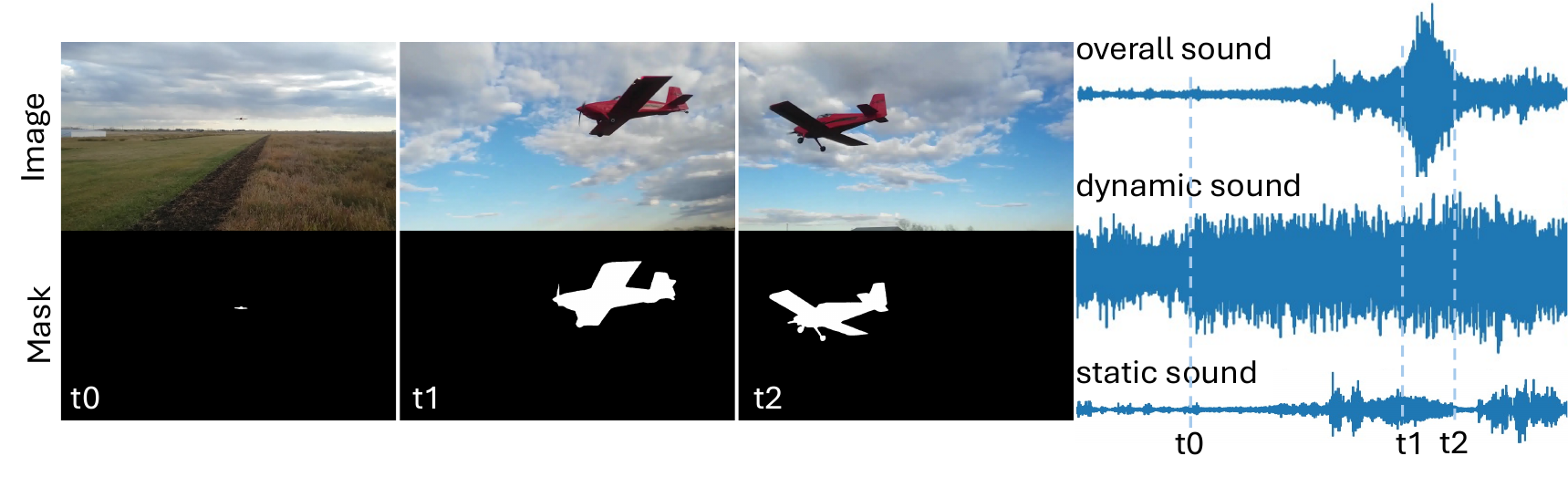}
    \caption{Dynamic source rendering on a plane fly-by clip: a time-varying 3D anchor from SAM3 \cite{sam3} and Depth Anything 3 \cite{da3} separates dynamic and static sound components, with the dynamic layer peaking near closest approach and attenuating with distance}
    \label{fig:dyn}
\end{figure}

\noindent\textbf{Dynamic sources.}
Our renderer also naturally supports dynamic sources when a time-varying 3D anchor is available. In a newly tested plane fly-by example, SAM3 \cite{sam3} and Depth Anything 3 \cite{da3} recover a source trajectory that we render into dynamic and static layers. The dynamic component peaks near the point of closest approach and attenuates with distance. Looking ahead, advances in 4D visual reconstruction may enable richer image-to-4D audiovisual scene generation (Fig.~\ref{fig:dyn}).

%% file: assets/tables/supp-ablation-.tex
\begin{table*}[t]
\vspace*{-1em}
\begin{small}
    \centering
    \begin{tabular}{lcccccccc}
    \toprule
    & \multicolumn{5}{c}{Spatial Metrics} & \multicolumn{3}{c}{Semantic Metrics}\\
    \cmidrule(lr){2-6}  \cmidrule(lr){7-9}
Method & $\Delta_{\text{abs}}\theta$ $\downarrow$ & $\Delta_{\text{abs}}\varphi$ $\downarrow$ & $\Delta_{\text{Anglular}}$ $\downarrow$ & CC $\uparrow$ & AUC $\uparrow$ & D-CLAP$_{\text{R}}$ $\uparrow$ & D-CLAP$_{\text{A}}$ $\uparrow$ & D-CLAP$_{\text{T}}$ $\uparrow$\\
\midrule
Ours (Point)      & 1.264 & 0.614 & 1.203 & 0.273 & 0.629 & 38.2\% & \textbf{0.521} & 0.419\\
Ours (w/o EQ)  & 0.815 & \textbf{0.196} & 0.814 & 0.602 & 0.806 & 58.8\% & 0.467 & 0.443 \\
Ours (No merge) & 0.800 & 0.359 & 0.843 & 0.586 & 0.807 & 39.7\% & 0.271 & 0.324 \\
Ours (All merge) & 1.032 & 0.331 & 1.012 & 0.445 & 0.719 & 45.6\% & 0.461 & 0.329 \\
Ours (Boundary perturb) & 0.760 & 0.228 & 0.791 & 0.627 & 0.824 & 69.1\% & 0.483 & 0.456 \\
Ours (Depth perturb) & 0.732 & 0.248 & 0.781 & 0.628 & 0.821 & 64.7\% & 0.484 & 0.450 \\
\midrule
Ours (Full) & \textbf{0.672} & 0.216 & \textbf{0.728} & \textbf{0.658} & \textbf{0.838} & \textbf{67.6\%} & {0.480} & \textbf{0.457} \\
    \bottomrule
    \end{tabular}
    \caption{\textbf{Ablation studies on \ourdataset.} We analyze the effect of source representation, equalization, grounding/merging strategy, and robustness to imperfect visual inputs. \textbf{Ours (Point)} replaces region-based sources with point sources. \textbf{Ours (w/o EQ)} removes the equalization module. \textbf{Ours (No merge)} treats tile-level masks independently, while \textbf{Ours (All merge)} merges all masks from the same category. \textbf{Ours (Boundary perturb)} and \textbf{Ours (Depth perturb)} evaluate robustness to noisy mask boundaries and depth estimates, respectively. Results show that each component contributes to the final performance, with our full model achieving the strongest overall spatial and semantic quality.}
    \label{tab:ablation}
    \end{small}
\end{table*}